\documentclass{article} 

\usepackage{iclr_arxiv,times}

\usepackage[utf8]{inputenc} 
\usepackage[T1]{fontenc}    
\usepackage{hyperref}       
\usepackage{url}            
\usepackage{booktabs}       
\usepackage{amsfonts}       
\usepackage{nicefrac}       
\usepackage{microtype}      
\usepackage{xcolor}         
\usepackage{bm}
\usepackage{wrapfig}
\usepackage{subcaption}
\usepackage{caption}

\usepackage{rotating}
\usepackage{booktabs}
\usepackage{multicol, multirow}

\usepackage{graphicx}

\usepackage{tabularray}
\UseTblrLibrary{booktabs}
\usepackage{algorithm}
\usepackage{algorithmic}

\usepackage{amsmath,amsfonts,bm}

\def\eqref#1{equation~\ref{#1}}

\def\1{\bm{1}}

\def\rmA{{\mathbf{A}}}

\def\rmO{{\mathbf{O}}}

\def\vx{{\bm{x}}}

\def\vz{{\bm{z}}}

\def\vepsilon{{\bm{\epsilon}}}

\def\mI{{\bm{I}}}

\DeclareMathAlphabet{\mathsfit}{\encodingdefault}{\sfdefault}{m}{sl}
\SetMathAlphabet{\mathsfit}{bold}{\encodingdefault}{\sfdefault}{bx}{n}

\def\gD{{\mathcal{D}}}

\def\gN{{\mathcal{N}}}

\def\gU{{\mathcal{U}}}

\newcommand{\E}{\mathbb{E}}

\usepackage{amsmath}
\usepackage{amssymb}
\usepackage{mathtools}
\usepackage{amsthm}
\usepackage{float}
\usepackage{pifont}
\usepackage{booktabs}
\usepackage{xspace}

\usepackage[capitalize,noabbrev]{cleveref}

\newcommand{\ours}{{OneDP}\xspace}

\title{One-Step Diffusion Policy: Fast Visuomotor Policies via Diffusion Distillation}

\author{Zhendong Wang$^{1,2}$\thanks{Work done during an internship at NVIDIA}, Zhaoshuo Li$^{1}$, Ajay Mandlekar$^{1}$, Zhenjia Xu$^{1}$, Jiaojiao Fan$^{1}$, \\
    \textbf{Yashraj Narang$^{1}$, Linxi Fan$^{1}$, Yuke Zhu$^{1,2}$, Yogesh Balaji$^{1}$, Mingyuan Zhou$^{2}$,} \\
    \textbf{Ming-Yu Liu$^{1}$, Yu Zeng$^{1}$} \\
    $^1$NVIDIA, $^2$The University of Texas at Austin
}

\iclrfinalcopy 
\begin{document}

\maketitle

\begin{abstract}

Diffusion models, praised for their success in generative tasks, are increasingly being applied to robotics, demonstrating exceptional performance in behavior cloning. However, their slow generation process stemming from iterative denoising steps poses a challenge for real-time applications in resource-constrained robotics setups and dynamically changing environments.
In this paper, we introduce the One-Step Diffusion Policy (\ours), a novel approach that distills knowledge from pre-trained diffusion policies into a single-step action generator, significantly accelerating response times for robotic control tasks. We ensure the distilled generator closely aligns with the original policy distribution by minimizing the Kullback-Leibler (KL) divergence along the diffusion chain, requiring only $2\%$-$10\%$ additional pre-training cost for convergence. We evaluated \ours on 6 challenging simulation tasks as well as 4 self-designed real-world tasks using the Franka robot. The results demonstrate that \ours not only achieves state-of-the-art success rates but also delivers an order-of-magnitude improvement in inference speed, boosting action prediction frequency from 1.5 Hz to 62 Hz, establishing its potential for dynamic and computationally constrained robotic applications. We share the project page here \href{https://research.nvidia.com/labs/dir/onedp/}{https://research.nvidia.com/labs/dir/onedp/}. 
\end{abstract}

\begin{figure}[ht]
    \centering
    \includegraphics[width=0.97\textwidth]{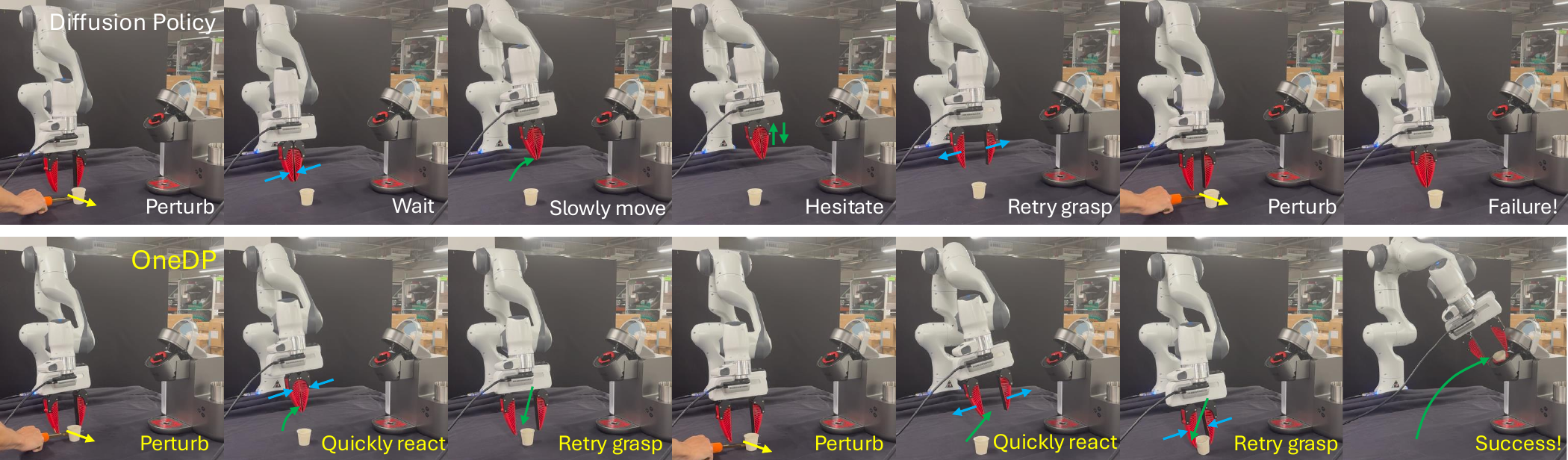} \\
    \includegraphics[height=3.3cm]{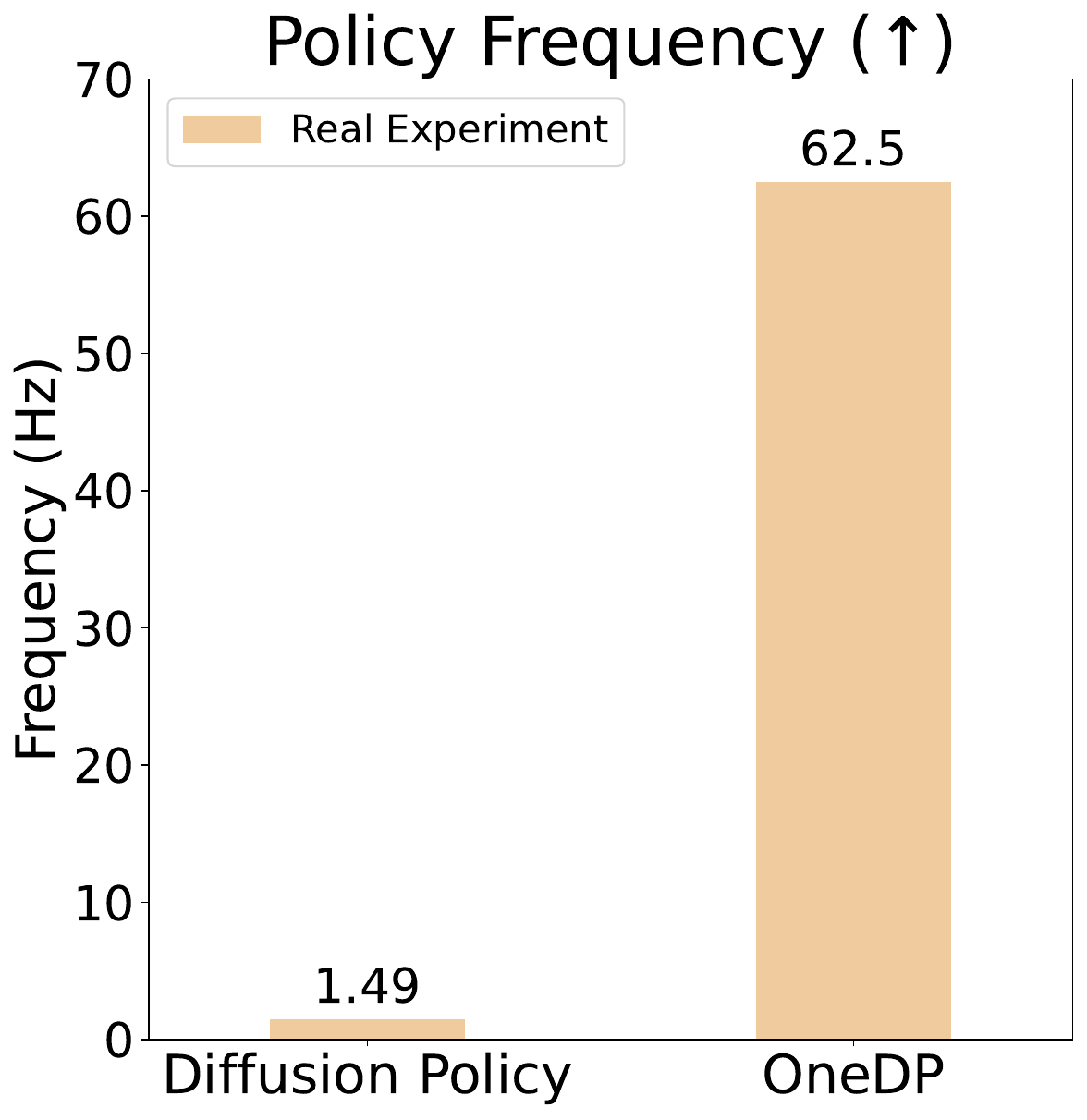}
    \includegraphics[height=3.3cm]{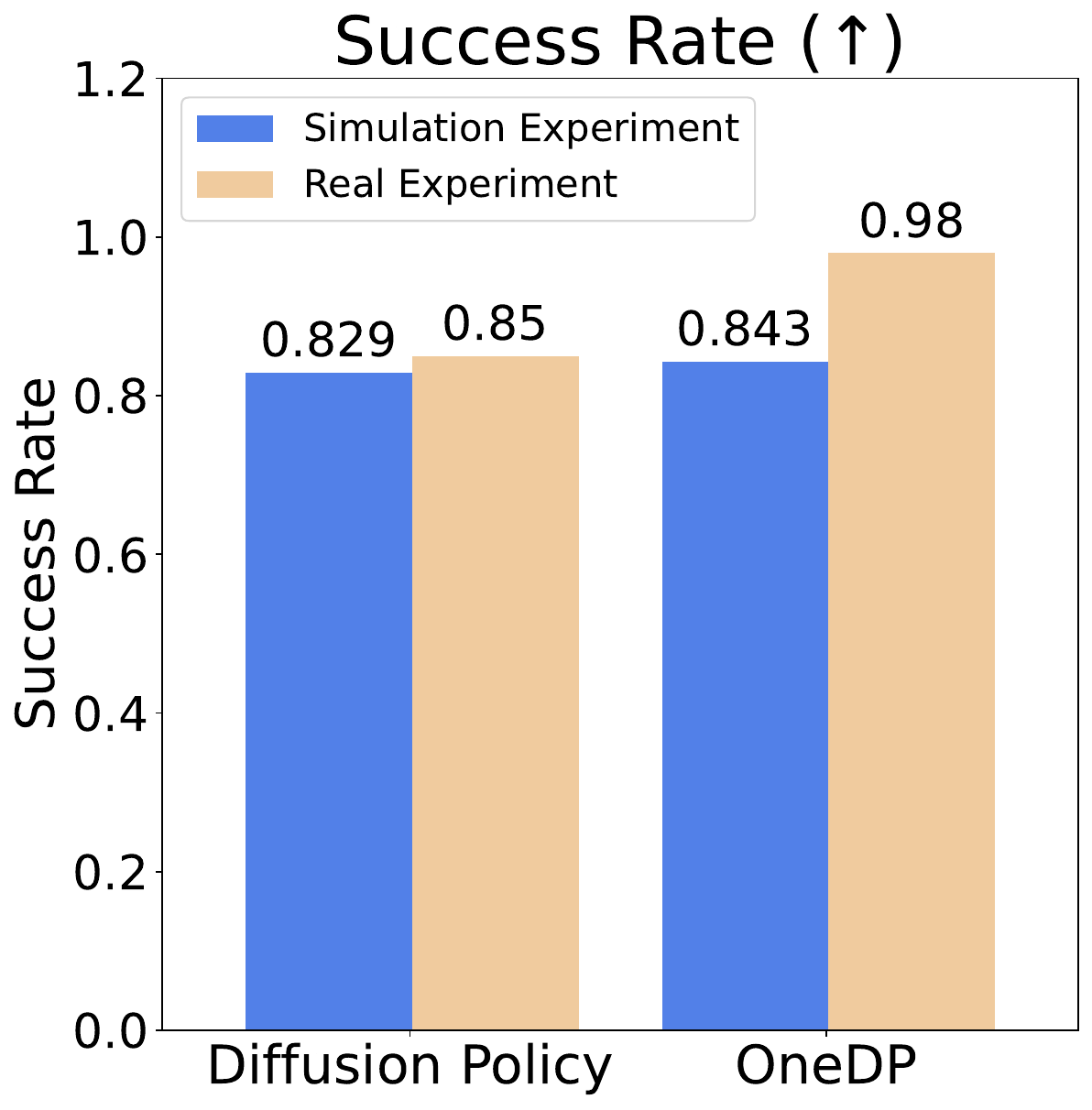}
    \includegraphics[height=3.3cm]{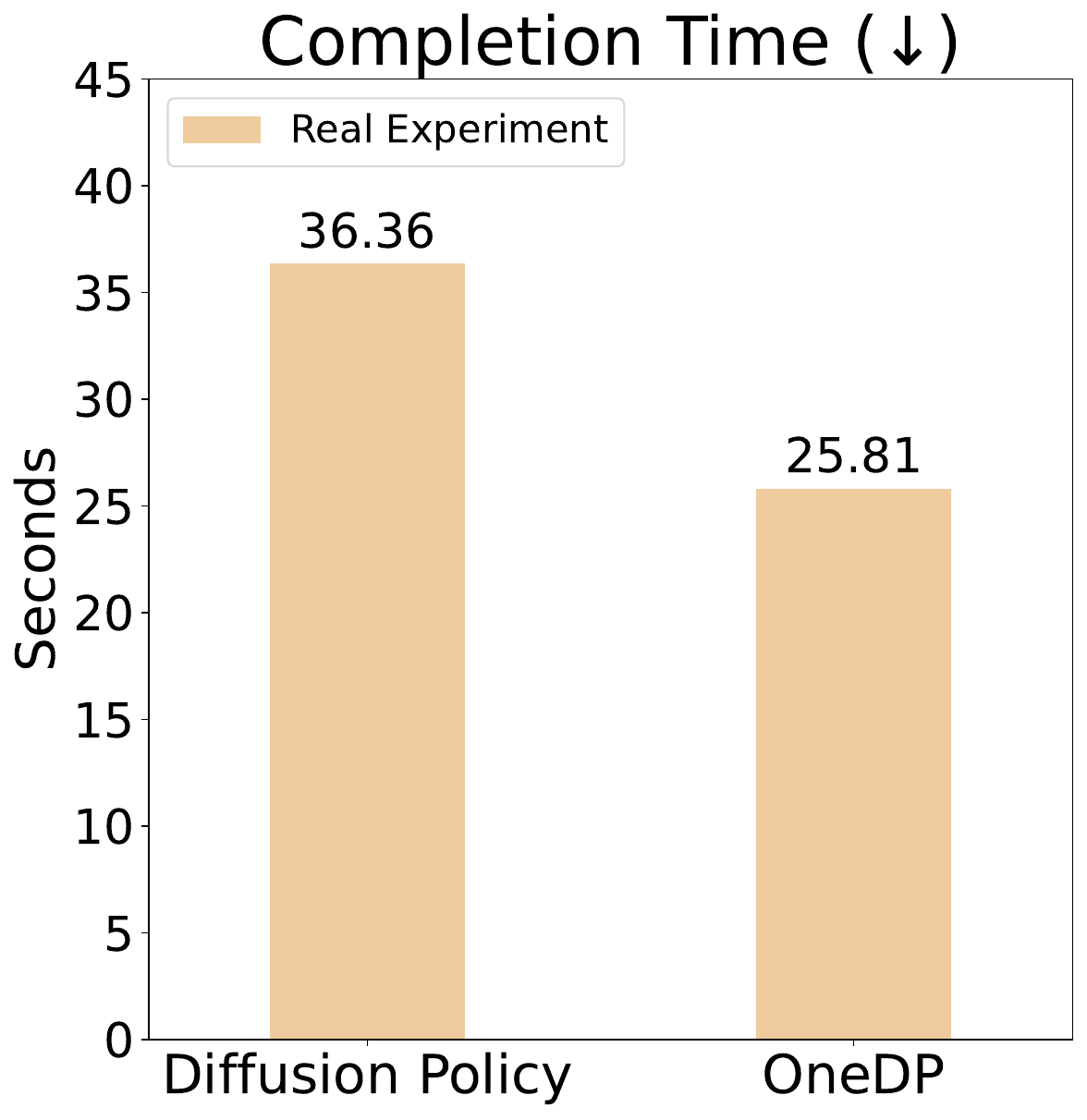}
    \includegraphics[height=3.3cm]{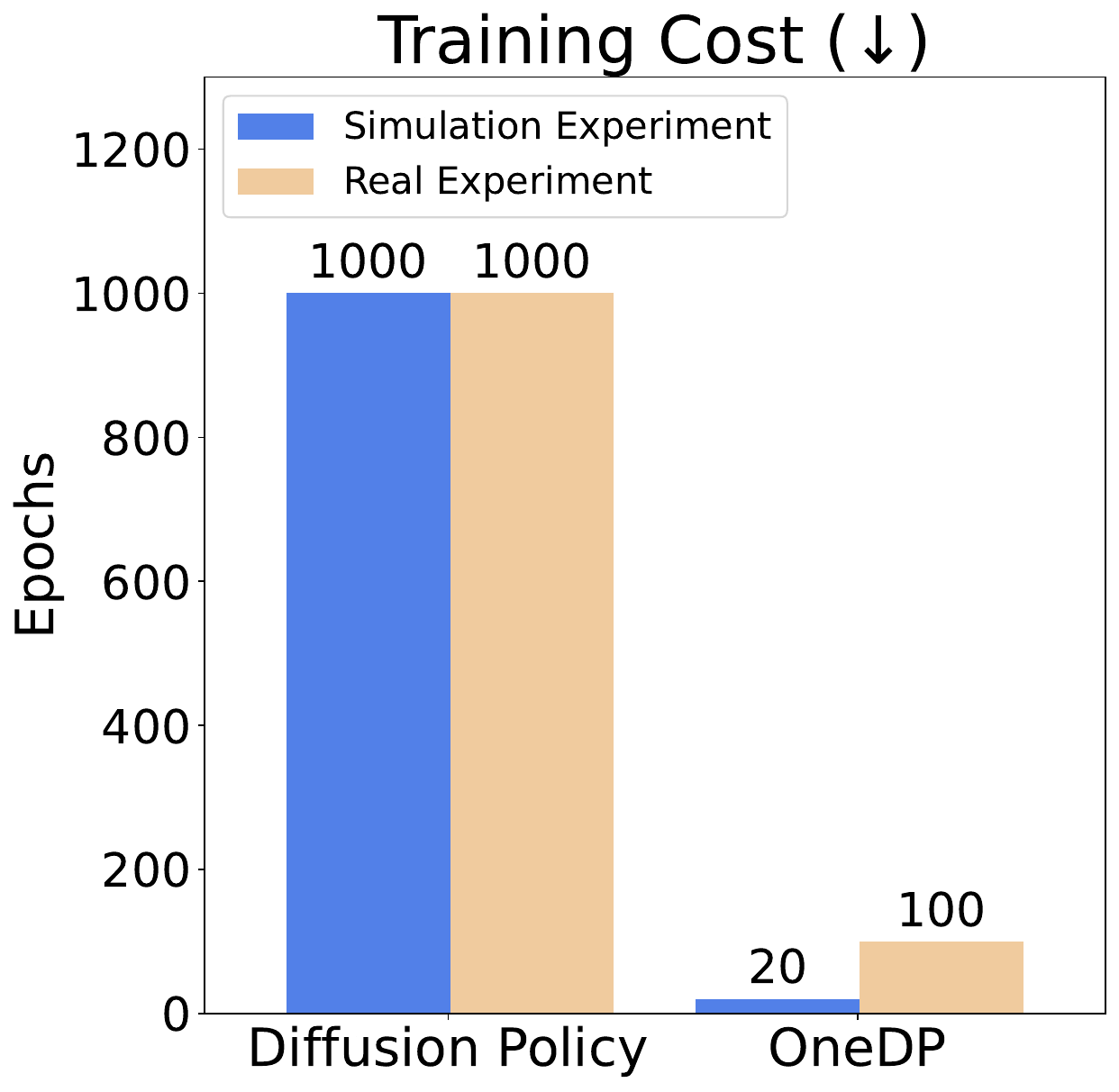}
    \caption{\textbf{Comparison of Diffusion Policy and One-Step Diffusion Policy (\ours).} We demonstrate the rapid response of \ours to changes in dynamic environments through real-world experiments. The first row illustrates how Diffusion Policy \citep{chi2023diffusion} struggles to adapt to environment changes (here, object perturbation) and fails to complete the task due to its slow inference speed. In contrast, the second row highlights \ours's quick and effective response. The third row offers a quantitative comparison: in the first panel, \ours executes action prediction much faster than  Diffusion Policy. This enhanced responsiveness results in a higher average success rate across multiple tasks, particularly in real-world scenarios, as depicted in the second panel. The third panel reveals that \ours also completes tasks more swiftly. The final panel indicates that distillation of  \ours requires only a small fraction of the pre-training cost.}
    \vspace{-3mm}
    \label{fig:teaser1}
\end{figure}

\section{Introduction}

Diffusion models \citep{sohl2015deep,ho2020denoising} have emerged as a leading approach to generative AI, achieving remarkable success in diverse applications such as text-to-image generation \citep{saharia2022photorealistic, ramesh2022hierarchical, rombach2022high}, video generation \citep{ho2022video, openai_sora}, and online/offline reinforcement learning (RL) \citep{wang2022diffusion, chen2023score, hansen2023idql, psenka2023learning}. Recently, \citet{chi2023diffusion,team2024octo,reuss2023goal,ze20243d,ke20243d, prasad2024consistency} demonstrated impressive results of diffusion models in imitation learning for robot control. In particular, \citet{chi2023diffusion} introduces the diffusion policy and achieves a state-of-the-art imitation learning performance on a variety of robotics simulation and real-world tasks.

However, because of the necessity of traversing the reverse diffusion chain, the slow generation process of diffusion models presents significant limitations for their application in robotic tasks. This process involves multiple iterations to pass through the same denoising network, potentially thousands of times \citep{song2020denoising, wang2023diffusiongan}. Such a long inference time restricts the practicality of using the diffusion policy \citep{chi2023diffusion}, which by default runs at $1.49$ Hz, in scenarios where quick response and low computational demands are essential. While classical tasks like block stacking or part assembly may accommodate slower inference rates, more dynamic activities involving human interference or changing environments require quicker control responses \citep{prasad2024consistency}. 
In this paper, we aim to significantly reduce inference time through diffusion distillation and achieve responsive robot control.

Considerable research has focused on streamlining the reverse diffusion process for image generation, aiming to complete the task in fewer steps.
A prominent approach interprets diffusion models using stochastic differential equations (SDE) or ordinary differential equations (ODE) and employs advanced numerical solvers for SDE/ODE to speed up the process \citep{song2020denoising, liu2022pseudo, karras2022elucidating, lu2022dpmsolver}. Another avenue explores distilling diffusion models into generators that require only one or a few steps through Kullback-Leibler (KL) optimization or adversarial training \citep{salimans2022progressive, song2023consistency, luo2024diff, yin2024one}. However, accelerating diffusion policies for robotic control has been largely underexplored. Consistency Policy \citep{prasad2024consistency} (CP) employs the consistency trajectory model (CTM) \citep{kim2023consistency} to adapt the pre-trained diffusion policy into a few-step CTM action generator. Despite this, several iterations for sampling are still required to maintain good empirical performance.

In this paper, we introduce the One-Step Diffusion Policy (\ours), which distills knowledge from pre-trained diffusion policies into a one-step diffusion-based action generator, thus maximizing inference efficiency through a single neural network feedforward operation. We demonstrate superior results over baselines in \Cref{fig:teaser1}.
Inspired by the success of SDS \citep{poole2022dreamfusion} and VSD \citep{wang2024prolificdreamer} in text-to-3D generation, we propose a policy-matching distillation method for robotic control. The training of \ours consists of three key components: a one-step action generator, a generator score network, and a pre-trained diffusion-policy score network. To align the generator distribution with the pre-trained policy distribution, we minimize the KL divergence over diffused actions produced by the generator, with the gradient of the KL expressed as a score difference loss.
By initializing the action generator and the generator score network with the identical pre-trained model, our method not only preserves or enhances the performance of the original model, but also requires only $2\%$-$10\%$ additional pre-training cost for the distillation to converge. We compare our method with CP and demonstrate that it outperforms CP with a higher success rate across tasks, leveraging a single-step action generator and achieving 20$\times$ faster convergence. A detailed comparison with this approach is provided in \Cref{sec:related_work,sec:experiments}.

We evaluate our method in both simulated and real-world environments. In simulated experiments, we test \ours on the six most challenging tasks of the Robomimic benchmark \citep{mandlekar2021matters}. For real-world experiments, we design four tasks with increasing difficulty and deploy \ours on a Franka robot arm. In both settings, \ours demonstrated state-of-the-art success rates with single-step generation, performing $42\times$ faster in inference.



\section{One-Step Diffusion Policy}
\label{sec:methodology}

\subsection{Preliminaries}

Diffusion models are powerful generative models applied across various domains \citep{ho2020denoising, sohl2015deep, song2020score}. They function by defining a forward diffusion process that gradually corrupts the data distribution into a known noise distribution. Given a data distribution $p(\vx)$, the forward process adds Gaussian noise to samples, $\vx^0 \sim p(\vx)$, with each step defined as $\vx^k = \alpha_k \vx^0 + \sigma_k \vepsilon_k$, where $\vepsilon_k \sim \gN(\bm{0}, \mI)$. The parameters $\alpha_k$ and $\sigma_k$ are manually designed and vary according to different noise scheduling strategies.

A probabilistic model $p_{\theta}(\vx^{k-1} | \vx^k)$ is then trained to reverse this diffusion process, enabling data generation from pure noise. DDPM \citep{ho2020denoising} uses discrete-time scheduling with a noise-prediction model $\epsilon_{\theta}$ to parameterize $p_{\theta}$, while EDM \citep{karras2022elucidating} employs continuous-time diffusion with $\vx^0$-prediction. We use epsilon prediction $\epsilon_\theta$ in our derivation. The diffusion model is trained using the denoising score matching loss \citep{ho2020denoising,song2020score}.

Once trained, we can estimate the unknown score $s(\vx^k)$ at a diffused sample $\vx^k$ as:
\begin{equation} \label{eq:score_estimate}
    s(\vx^k) = - \frac{\epsilon^*(\vx^k, k)}{\sigma_k} \approx - \frac{\epsilon_\theta(\vx^k, k)}{\sigma_k},
\end{equation}
where $\epsilon^*(\vx^k, k)$ is the true noise added at time $k$ and we denote $s_\theta(\vx^k) = - \frac{\epsilon_\theta(\vx^k, k)}{\sigma_k}$. With a score estimate, clean data $\vx^0$ can be sampled by reversing the diffusion chain \citep{song2020score}. This requires multiple iterations through the estimated score network, making it inherently slow.

\citet{wang2022diffusion, chi2023diffusion} extend diffusion models as expressive and powerful policies for offline RL and robotics. In robotics, a set of past observation images, $\rmO$, is used as input to the policy. An action chunk, $\rmA$, which consists of a sequence of consecutive actions, forms the output of the policy. Diffusion policy is represented as a conditional diffusion-based action prediction model, 
\begin{equation} \label{eq:diffusion_policy}
    \pi_\theta(\rmA^0 | \rmO) := \int \cdots \int \gN(\rmA^K; \bm{0}, \mI) \prod_{k=K}^{k=1} p_{\theta}(\rmA^{k-1}| \rmA^k, \rmO) d\rmA^{K} \cdots d\rmA^{1},
\end{equation}
The explicit form of $\pi_\theta(\rmA^0 | \rmO)$ is often impractical due to the complexity of integrating actions from $\rmA^K$ to $\rmA^1$. However, we can obtain action chunk samples from it by iterative denoising. More details are provided in \Cref{sec:appendix_prelim}

\subsection{One-Step Diffusion Policy} \label{sec:diff_distill}

\begin{figure}[t]
    \centering
    \includegraphics[width=\linewidth]{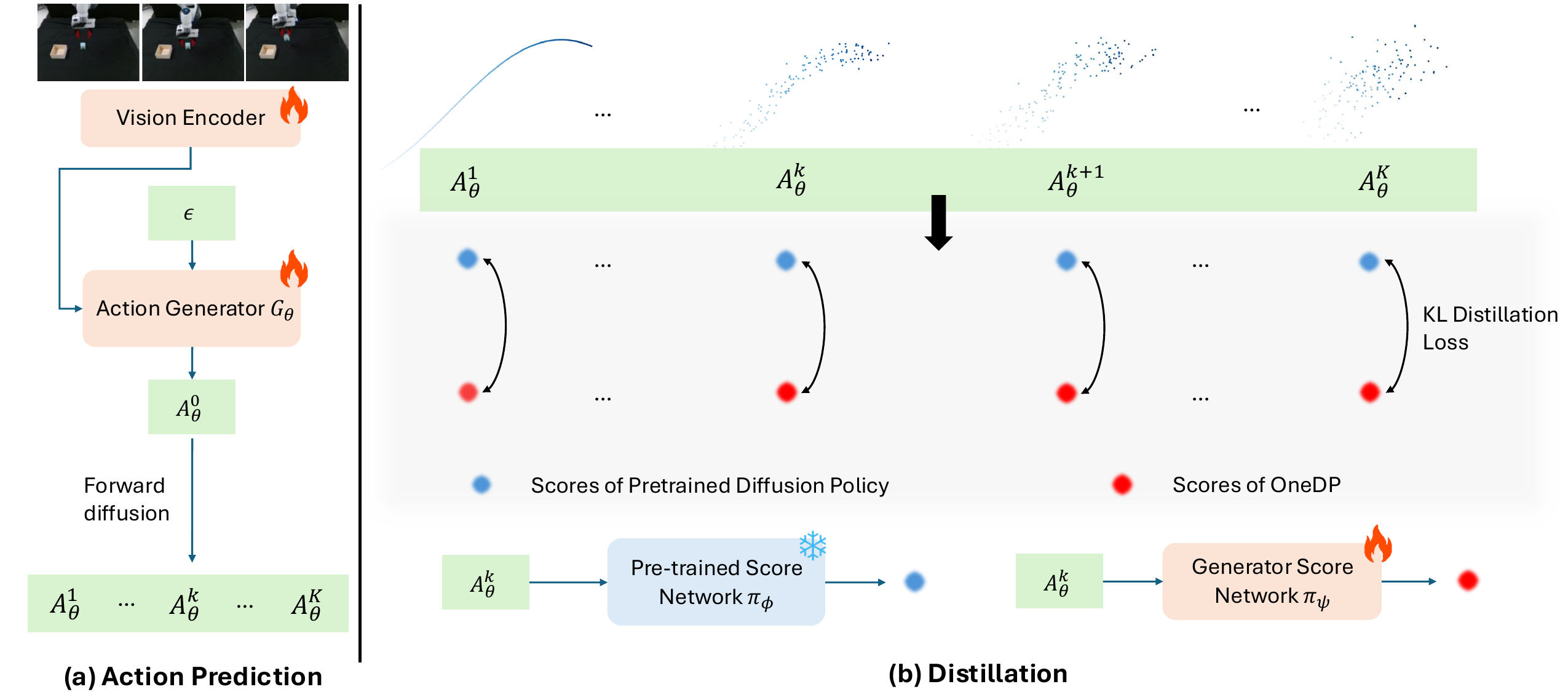}
    \caption{\textbf{Diffusion Distillation Pipeline.} a) Our one-step action generator processes image-based visual observations alongside a random noise input to deliver single-step action predictions. b) We implement KL-based distillation across the entire forward diffusion chain. Direct computation of the KL divergence is often impractical; however, we can effectively utilize the gradient of the KL, formulated into a score-difference loss. The pre-trained score network $\pi_\phi$ remains fixed while the action generator $G_\theta$ and the generator score network $\pi_\psi$ are trained.}
    \label{fig:pipeline}
\end{figure}

Action sampling through the vanilla diffusion policies is notoriously slow due to the need of tens to hundreds of iterative inference steps. 
The latency issue is critical for computationally sensitive robotic tasks or tasks that require high control frequency. 
Although employing advanced ODE solvers \citep{song2020denoising,karras2022elucidating} could help speed up the sampling procedure, empirically at least ten iterative steps are required to ensure reasonable performance. Here, we introduce a training-based diffusion policy distillation method, which distills the knowledge of a pre-trained diffusion policy into a single-step action generator, enabling fast action sampling. 

We propose a one-step implicit action generator $G_\theta$, from which actions can be easily obtained as follows,
\begin{equation} \label{eq:osdp_sample}
    \vz \sim \gN(\bm{0}, \mI), \rmA_\theta = G_\theta(\vz, \rmO).
\end{equation}
We define the action distribution generated by $G_\theta$ as $p_{G_\theta}$. Assuming the existence of a pre-trained diffusion policy $\pi_\phi(\rmA | \rmO)$ defined by \Cref{eq:diffusion_policy} and parameterized by $\epsilon_\phi$, its corresponding action distribution is denoted as $p_{\pi_\phi}$. Drawing inspiration from the success of SDS \citep{poole2022dreamfusion} and VSD \citep{wang2024prolificdreamer} in text-to-3D applications, we propose using the following reverse KL divergence to align the distributions $p_{G_\theta}$ and $p_{\pi_\phi}$,
\begin{equation*}
    \gD_{KL} (p_{G_\theta} || p_{\pi_\phi}) = \E_{\vz \sim \gN(\bm{0}, \mI),\rmA_\theta = G_\theta(\vz, \rmO)} \left[ \log p_{G_\theta}(\rmA_\theta | \rmO) - \log p_{\pi_\phi}(\rmA_\theta | \rmO) \right].
\end{equation*}
It is generally intractable to estimate this loss by directly computing the probability densities, since $p_{G_\theta}$ is an implicit distribution and $p_{\pi_\phi}$ involves integrals that are impractical (\Cref{eq:diffusion_policy}). However, we only need the gradient with respect to $\theta$ to train our generator by gradient descent: 
\begin{equation} \label{eq:kl_loss}
    \nabla_{\theta} \gD_{KL} (p_{G_\theta} || p_{\pi_\phi}) = \E_{\scriptstyle \substack{\vz \sim \gN(\bm{0}, \mI), \\ \rmA_\theta = G_\theta(\vz, \rmO)}} \left[ (\nabla_{\rmA_\theta}\log p_{G_\theta}(\rmA_\theta | \rmO) - \nabla_{\rmA_\theta}\log p_{\pi_\phi}(\rmA_\theta | \rmO)) \nabla_{\theta} \rmA_\theta \right].
\end{equation}
Here $s_{p_{G_\theta}}(\rmA_\theta) = \nabla_{\rmA_\theta}\log p_{G_\theta}(\rmA_\theta | \rmO)$ and $s_{p_{\pi_\phi}}(\rmA_\theta) = \nabla_{\rmA_\theta}\log p_{\pi_\phi}(\rmA_\theta | \rmO)$ are the scores of the $p_{G_\theta}$ and $p_{\pi_\phi}$ respectively. Computing this gradient still presents two significant challenges: First, the scores tend to diverge for samples from $p_{G_\theta}$ that have a low probability in $p_{\pi_\phi}$, especially when $p_{\pi_\phi}$ may approach zero. Second, the primary tool for estimating these scores, the diffusion models, only provides scores for the diffused distribution. 

Inspired by Diffusion-GAN \citep{wang2023diffusiongan}, which proposed to optimize statistical divergence, such as the Jensen–Shannon divergence (JSD), throughout diffused data samples, we propose to similarly optimize the KL divergence outlined in \Cref{eq:kl_loss} across diffused action samples as described below:
\begin{equation} \label{eq:ekl_loss_vsd}
    \nabla_{\theta} \E_{k\sim\gU} [\gD_{KL} (p_{G_\theta,k} || p_{\pi_\phi,k})] = \E_{\scriptstyle \substack{\vz \sim \gN(\bm{0}, \mI), k\sim\gU \\ \rmA_\theta = G_\theta(\vz, \rmO) \\ \rmA_{\theta}^k \sim q(\rmA_{\theta}^k | \rmA_{\theta}, k)}} \left[ w(k) (s_{p_{G_\theta}}(\rmA_{\theta}^k) - s_{p_{\pi_\phi}}(\rmA_{\theta}^k)) \nabla_{\theta} \rmA_{\theta}^k \right].
\end{equation}
where $w(k)$ is a reweighting function, $q$ is the forward diffusion process and $s_{p_{\pi_\phi}}(\rmA_{\theta}^k)$ could be obtained through \Cref{eq:score_estimate} with $\epsilon_\phi$. 
In order to estimate the score of the generator distribution, $s_{p_{G_\theta}}$, we introduce an auxiliary diffusion network $\pi_\psi(\rmA | \rmO)$, parameterized by $\epsilon_\psi$. 
We follow the typical way of training diffusion policies, which optimizes $\psi$ by treating $p_{G_\theta}$ as the target action distribution \citep{wang2024prolificdreamer},
\begin{equation} \label{eq:psi_update}
    \min_\psi \E_{\vx^k \sim q(\vx^k|\vx^0), \vx^0 = \text{stop-grad}(G_\theta(\vz)), \vz \sim \gN(\bm{0}, \mI), k\sim \gU}[\lambda(k) \cdot || \epsilon_\psi(\vx^k, k) - \vepsilon_k||^2].
\end{equation}
Then we can obtain $s_{p_{\pi_\psi}}(\rmA_{\theta}^k)$ by applying $\epsilon_\psi$ to \Cref{eq:score_estimate}. We approximate $s_{p_{G_\theta}}(\rmA_{\theta}^k)$ in \Cref{eq:ekl_loss_vsd} with $s_{p_{\pi_\psi}}(\rmA_{\theta}^k)$. We iteratively update the generator parameters $\theta$ by \Cref{eq:ekl_loss_vsd}, and the generator score network parameter $\psi$ by \Cref{eq:psi_update}. The parameter of the prertrained diffusion policy $\phi$ is fixed throughout the training. During inference, we directly perform one-step sampling with \Cref{eq:osdp_sample}. We name our algorithm \ours-S, where \textit{S} denotes the stochastic policy. 

When we apply a deterministic action generator by omitting random noise $\vz$, such that $\rmA_\theta = G_\theta(\rmO)$, the distribution $p_{G_\theta}$ becomes a Dirac delta function centered at $G_\theta(\rmO)$, that is, $p_{G_\theta} = \delta_{G_\theta(\rmO)}(\rmA)$. Consequently, $s_{p_{G_\theta}}(\rmA_{\theta}^k)$ can be explicitly solved as follows:
\begin{equation} \label{eq:score_sds}
    s_{p_{G_\theta}}(\rmA_{\theta}^k) = \nabla_{\rmA_{\theta}^k} \log p_\theta(\rmA_{\theta}^k) = \nabla_{\rmA_{\theta}^k} \log p_\theta(\rmA_{\theta}^k | \rmA_{\theta}) = - \frac{\vepsilon_k}{\sigma_k}; \rmA_{\theta}^k = \alpha_k \rmA_{\theta} + \sigma_k \vepsilon_k, \vepsilon_k \sim \gN(\bm{0}, \mI).
\end{equation}
By incorporating \Cref{eq:score_sds} into \Cref{eq:ekl_loss_vsd}, we can have a simplified loss function without the need of introducing the generator score network:
\begin{equation} \label{eq:ekl_loss_sds}
    \nabla_{\theta} \E_{k\sim\gU} [\gD_{KL} (p_{G_\theta,k} || p_{\pi_\phi,k})] = \E_{\scriptstyle \substack{\vz \sim \gN(\bm{0}, \mI), k\sim\gU \\ \rmA_\theta = G_\theta(\vz, \rmO) \\ \rmA_{\theta}^k \sim q(\rmA_{\theta}^k | \rmA_{\theta}, k)}} \left[ \frac{w(k)}{\sigma_k} (\epsilon_\phi(\rmA_{\theta}^k, k)) - \epsilon_k) \nabla_{\theta} \rmA_{\theta}^k \right].
\end{equation}
We name this deterministic diffusion policy distillation \ours -D. We illutrate our training pipeline in \Cref{fig:pipeline}, and summarize our algorithm training in \Cref{alg:onedp_training}.

\textbf{Policy Discussion.} A stochastic policy, which encompasses deterministic policies, is more versatile and better suited to scenarios requiring exploration, potentially leading to better convergence at a global optimum \citep{haarnoja2018soft}. In our case, \ours-D simplifies the training process, though it may exhibit slightly weaker empirical performance. We offer a comprehensive comparison between \ours-S and \ours-D in \Cref{sec:experiments}.

\begin{minipage}{0.46\textwidth}
    \textbf{Distillation Discussion.} We discuss the benefits of optimizing the expectational reverse KL divergence. First, reverse KL divergence typically induces mode-seeking behavior, which has been shown to improve empirical performance in offline RL \citep{chen2023score}. Therefore, we anticipate that reverse KL-based distillation offers similar advantages for robotic tasks. Second, as demonstrated by \citet{wang2023diffusiongan}, optimizing JSD, a combination of KLs, between diffused action samples provides stronger performance when dealing with distributions with misaligned supports. This aligns with our approach of performing KL optimization over the diffused distribution. 
\end{minipage} \hfill
\begin{minipage}{0.5\textwidth}
    \begin{algorithm}[H]
    \caption{OneDP Training}
    \label{alg:onedp_training}
    \begin{algorithmic}[1]
    \STATE \textbf{Inputs: } action generator $G_\theta$, generator score network $\pi_\psi$, pre-trained diffusion policy $\pi_\phi$.
    \STATE \textbf{Initializaiton} $G_\theta \leftarrow \pi_\phi$, $\pi_\psi \leftarrow \pi_\phi$.
    \WHILE{not converged}
        \STATE Sample $\rmA_\theta = G_\theta(\vz, \rmO), \vz \sim \gN(\mathbf{0}, \mI)$.
        \STATE Diffuse 
        $\rmA_{\theta}^k = \alpha_k \rmA_{\theta} + \sigma_k \vepsilon_k, \vepsilon_k \sim \gN(\bm{0}, \mI)$.
        \IF{OneDP-S}
        \STATE Update $\psi$ by \Cref{eq:psi_update}
        \STATE Update $\theta$ by \Cref{eq:ekl_loss_vsd}
        \ELSIF{OneDP-D}
        \STATE Update $\theta$ by \Cref{eq:ekl_loss_sds}
        \ENDIF
    \ENDWHILE
    \end{algorithmic}
    \end{algorithm}
\end{minipage}

\subsection{Implementation Details}

\textbf{Diffusion Policy.} Following \citet{chi2023diffusion}, we construct a diffusion policy using a 1D temporal convolutional neural network (CNN) \citep{janner2022planning} based U-Net and a standard ResNet18 (without pre-training) \citep{he2016deep} as the vision encoder. We implement the diffusion policy with two noise scheduling methods: DDPM \citep{ho2020denoising} and EDM \citep{karras2022elucidating}. We use $\epsilon$ noise prediction for discrete-time (100 steps) diffusion and $x^0$ prediction for continuous-time diffusion, respectively. The EDM scheduling is essential for Consistency Policy \citep{prasad2024consistency} due to the use of CTM \citep{kim2023consistency}. For DDPM, we set $\lambda(k)=1$ and use the original SDE and DDIM \citep{song2020denoising} sampling. 
For EDM, we use the default $\lambda(k) = \frac{\sigma_k^2 + \sigma_d^2}{(\sigma_k\sigma_d)^2}$ with $\sigma_d=0.5$. 
We use the second-order EDM sampler, which requires two neural network forwards per discretized step in the ODE. 

\textbf{Distillation.} We warm-start both the stochastic and deterministic action generator $G_\theta$, and the generator score network, $\epsilon_\psi$, by duplicating the neural-network structure and weights from the pre-trained diffusion policy, aligning with strategies from \citet{luo2024diff, yin2024one, xu2024ufogen}. Following DreamFusion \citep{poole2022dreamfusion}, we set $w(k) = \sigma_k^2$. In the discrete-time domain, distillation occurs over [2, 95] diffusion timesteps to avoid edge cases. In continuous-time, we employ the same log-normal noise scheduling as EDM \citep{karras2022elucidating} used during distillation. The generators operate at a learning rate of $1 \times 10^{-6}$, while the generator score network is accelerated to a learning rate of $2 \times 10^{-5}$. Vision encoders are also actively trained during the distillation process.

\section{Experiments}
\label{sec:experiments}
We evaluate \ours on a wide variety of tasks in both simulated and real environments. In the following sections, we first report the evaluation results in simulation across six tasks that include different complexity levels. Then we demonstrate the results in the real environment by deploying \ours in the real world with a Franka robot arm for object pick-and-place tasks and a coffee-machine manipulation task.  We compare our method with the pre-trained backbone Diffusion Policy \citep{chi2023diffusion} (DP) and related distillation baseline Consistency Policy \citep{prasad2024consistency} (CP). We also report the ablation study results in \Cref{sec:exp_ablation} to present more detailed analyses on our method and discuss the effect of different design choices.

\subsection{Simulation experiments} \label{sec:simulated_exp}

\begin{figure}[t]
    \centering
\begin{minipage}{\textwidth}
    \centering
    \includegraphics[width=\linewidth]{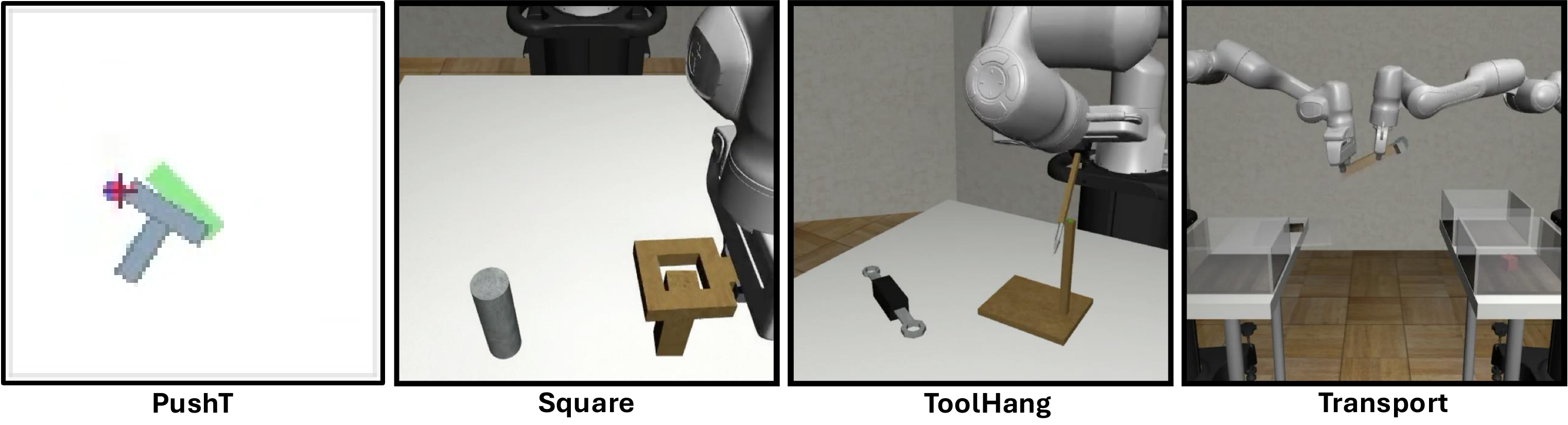}
    \caption{\textbf{Simulation tasks.} We evaluate our method against baselines on the single-robot tasks: PushT, Square, and ToolHang, as well as a dual-robot task Transport. Task difficulty increases from left to right. }
    \label{fig:sim_demo}
\end{minipage}
\begin{minipage}{\textwidth}
\centering
\captionof{table}{\textbf{Robomimic Benchmark Performance (Visual Policy) in DDPM}. We compare our proposed \ours-D and \ours-S, with DP under the default DDPM scheduling. We report the mean and standard deviation of success rates across 5 different training runs, each evaluated with 100 distinct environment initializations. Details of the evaluation procedure can be found in \Cref{sec:simulated_exp}. Our results demonstrate that \ours not only matches but can even outperform the pre-trained DP, achieving this with just one-step generation, resulting in an order of magnitude speed-up. }
\label{tab:main_result_ddpm}
\resizebox{\textwidth}{!}{
\begin{tabular}{c|cc|cccccc|c}
\hline
\textbf{Method} & \textbf{Epochs} & \textbf{NFE} & \textbf{PushT} & \textbf{Square-mh} & \textbf{Square-ph} & \textbf{ToolHang-ph} & \textbf{Transport-mh} & \textbf{Transport-ph} & \textbf{Avg} \\
\hline
{DP (DDPM)} & 1000 & 100 & \textbf{0.863 {\small $\pm$ 0.040}} & 0.846 {\small $\pm$ 0.023} & \textbf{0.926 {\small $\pm$ 0.023}} & 0.822 {\small $\pm$ 0.016} & 0.620{ \small $\pm$ 0.049} & 0.896 {\small $\pm$ 0.032} & 0.829 \\
\hline
\multirow{2}{*}{DP (DDIM)} & 1000 & 10 & 0.823{\small $\pm$ 0.023} & 0.850{\small $\pm$ 0.013} & 0.918{\small $\pm$ 0.009} & 0.828{\small $\pm$ 0.016} & 0.688{\small $\pm$ 0.020} & 0.908{\small $\pm$ 0.011} & 0.836 \\
 & 1000 & 1 & 0.000{\small $\pm$ 0.000} & 0.000{\small $\pm$ 0.000} & 0.000{\small $\pm$ 0.000} & 0.000{\small $\pm$ 0.000} & 0.000{\small $\pm$ 0.000} & 0.000{\small $\pm$ 0.000} & 0.000 \\
\hline
{\ours-D} & 20 & 1 & 0.802 {\small $\pm$ 0.057} & 0.846 {\small $\pm$ 0.028} & 0.926 {\small $\pm$ 0.011} & 0.808 {\small $\pm$ 0.046} & 0.676 {\small $\pm$ 0.029} & 0.896 {\small $\pm$ 0.013} & 0.826 \\
{\ours-S} & 20 & 1 & 0.816 {\small $\pm$ 0.058} & \textbf{0.864 {\small $\pm$ 0.042}} & \textbf{0.926 {\small $\pm$ 0.018}} & \textbf{0.850 {\small $\pm$ 0.033}} & \textbf{0.690 {\small$\pm$ 0.024}} & \textbf{0.914 {\small $\pm$ 0.021}} & \textbf{0.843} \\
\hline
\end{tabular}}
\centering
\captionof{table}{\textbf{Robomimic Benchmark Performance (Visual Policy) in EDM}. We compare our proposed \ours with CP under the EDM scheduling. EDM scheduling is required in CP to satisfy boundary conditions. We follow our evaluation metric and report similar values as in \Cref{tab:main_result_ddpm}. We also ablate Diffusion Policy with 1, 10 and 18 ODE steps, which utilizes 1, 19 and 35 NFE in EDM sampling. }
\label{tab:main_result_edm}
\resizebox{\textwidth}{!}{
\begin{tabular}{c|cc|cccccc|c}
\hline
\textbf{Method} & \textbf{Epochs} & \textbf{NFE} & \textbf{PushT} & \textbf{Square-mh} & \textbf{Square-ph} & \textbf{ToolHang-ph} & \textbf{Transport-mh} & \textbf{Transport-ph} & \textbf{Avg} \\
\hline
\multirow{3}{*}{DP (EDM)} & 1000 & 35 & \textbf{0.861{\small $\pm$ 0.030}} & 0.810{\small $\pm$ 0.026} & 0.898{\small $\pm$ 0.033} & \textbf{0.828{\small $\pm$ 0.019}} & 0.684{\small $\pm$ 0.019} & 0.890{\small $\pm$ 0.012} & 0.829 \\
 & 1000 & 19 & 0.851{\small $\pm$ 0.012} & \textbf{0.828{\small $\pm$ 0.015}} & 0.880{\small $\pm$ 0.014} & 0.794{\small $\pm$ 0.012} & 0.692{\small $\pm$ 0.009} & 0.860{\small $\pm$ 0.013} & 0.818 \\
 & 1000 & 1 & 0.000{\small $\pm$ 0.000} & 0.000{\small $\pm$ 0.000} & 0.000{\small $\pm$ 0.000} & 0.000{\small $\pm$ 0.000} & 0.000{\small $\pm$ 0.000} & 0.000{\small $\pm$ 0.000} & 0.000 \\
\hline
{CP} & 20 & 1 & 0.595{\small $\pm$ 0.141} & 0.120{\small $\pm$ 0.165} & 0.238{\small $\pm$ 0.219} & 0.238{\small $\pm$ 0.163} & 0.140{\small $\pm$ 0.148} & 0.174{\small $\pm$ 0.257} & 0.251 \\
{CP} & 450 & 1 & 0.828{\small $\pm$ 0.055} & 0.646{\small $\pm$ 0.047} & 0.776{\small $\pm$ 0.055} & 0.650{\small $\pm$ 0.046} & 0.378{\small $\pm$ 0.091} & 0.754{\small $\pm$ 0.120} & 0.672 \\
{CP} & 450 & 3 & 0.839{\small $\pm$ 0.037} & 0.710{\small $\pm$ 0.018} & 0.874{\small $\pm$ 0.022} & 0.626{\small $\pm$ 0.041} & 0.374{\small $\pm$ 0.051} & 0.848{\small $\pm$ 0.028} & 0.712 \\
\hline
{\ours-D} & 20 & 1 & 0.829{\small $\pm$ 0.052} & 0.776{\small $\pm$ 0.023} & 0.902{\small $\pm$ 0.040} & 0.762{\small $\pm$ 0.056} & 0.705{\small $\pm$ 0.038} & 0.898{\small $\pm$ 0.019} & 0.812 \\
{\ours-S} & 20 & 1 & {0.841{\small $\pm$ 0.042}} & 0.774{\small $\pm$ 0.033} & \textbf{0.910{\small $\pm$ 0.041}} & 0.824{\small $\pm$ 0.039} & \textbf{0.722{\small $\pm$ 0.025}} & \textbf{0.910{\small $\pm$ 0.027}} & \textbf{0.830} \\
\hline
\end{tabular}}
\vspace{-4mm}
\end{minipage}
\end{figure}

\textbf{Datasets.} \underline{Robomimic.} Proposed in \citep{mandlekar2021matters}, Robomimic is a large-scale benchmark for robotic manipulation tasks. The original benchmark consists of five tasks: Lift, Can, Square, Transport, and Tool Hang. We find that the the performance of state-of-the-art methods was already saturated on two easy tasks Lift and Can, and therefore only conduct the evaluation on the harder tasks Square, Transport and Tool Hang. For each of these tasks, the benchmark provides two variants of human demonstrations: proficient human (PH) demonstrations and mixed proficient/non-proficient human (MH) demonstrations. \underline{PushT.} Adapted from IBC \citep{florence2022implicit}, \citet{chi2023diffusion} introduced the PushT task, which involves pushing a T-shaped block into a fixed target using a circular end-effector. A dataset of 200 expert demonstrations is provided with RGB image observations.

\begin{figure}[ht]
    \centering
    \includegraphics[width=0.32\linewidth]{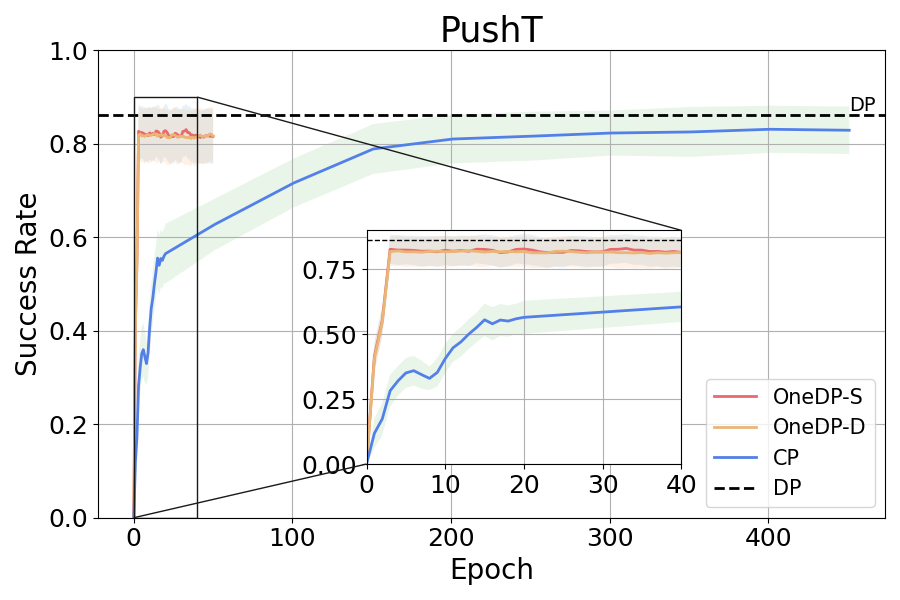}
    \includegraphics[width=0.32\linewidth]{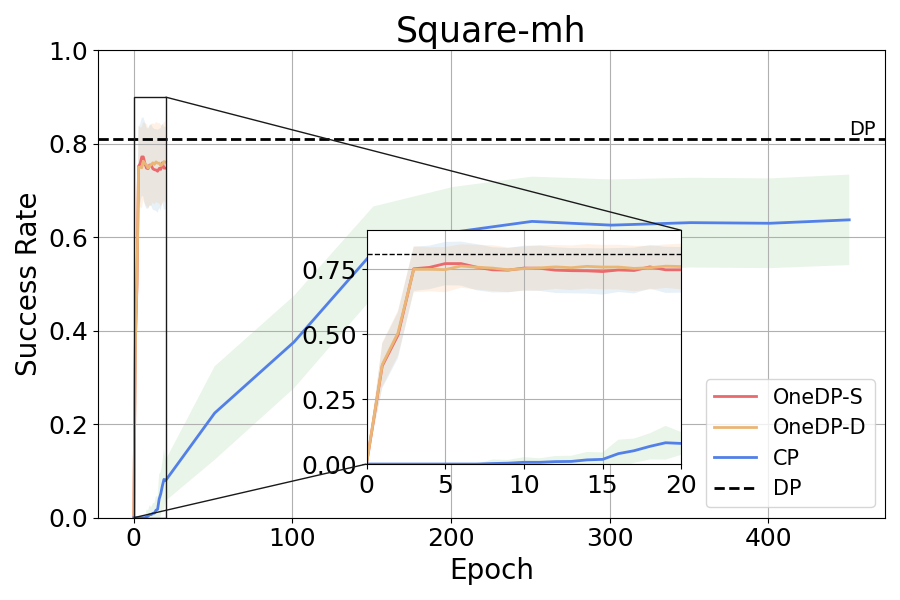}
    \includegraphics[width=0.32\linewidth]{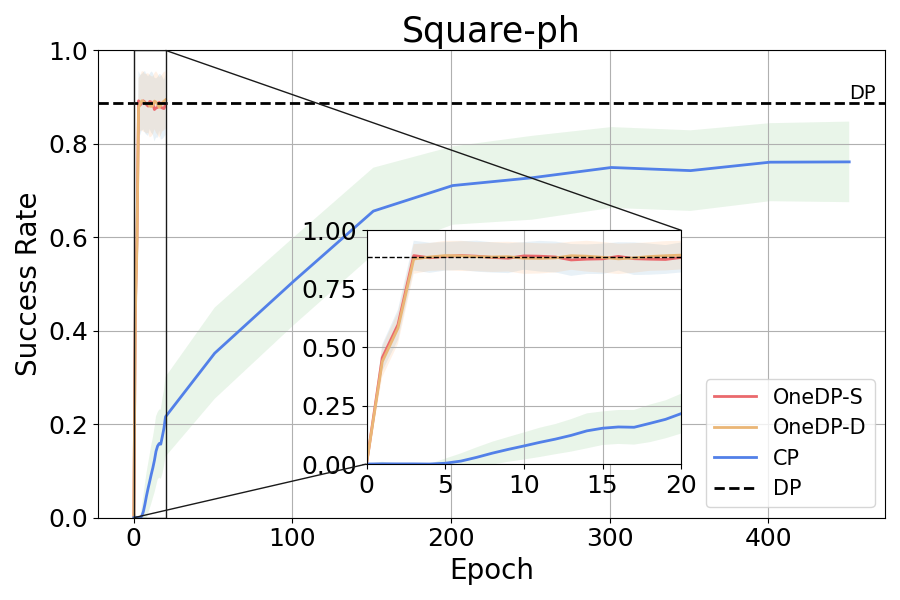} \\
    \includegraphics[width=0.32\linewidth]{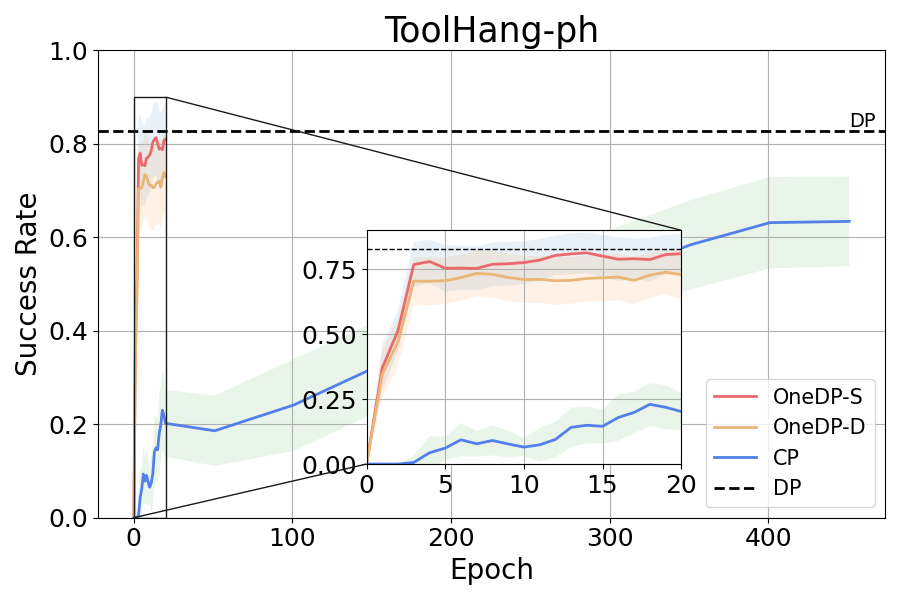}
    \includegraphics[width=0.32\linewidth]{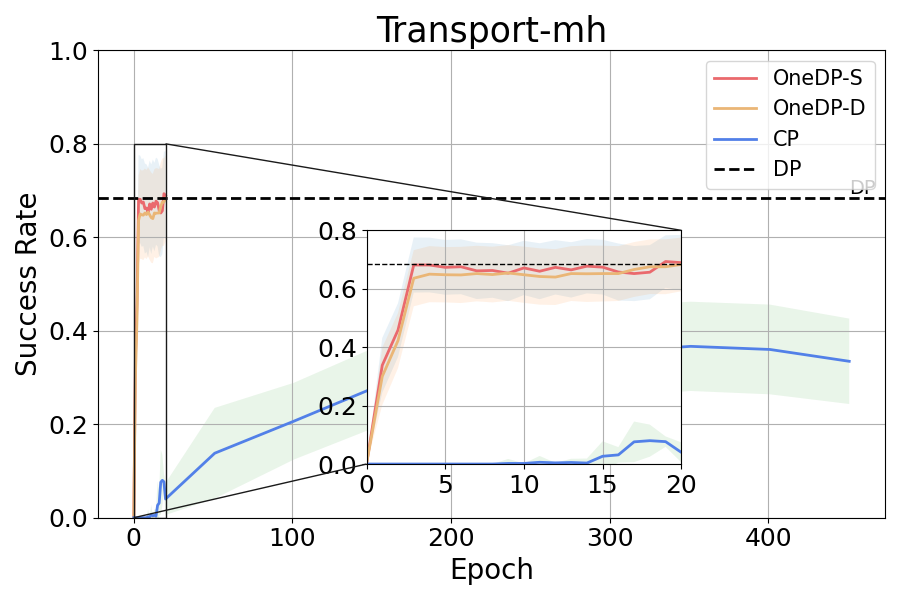}
    \includegraphics[width=0.32\linewidth]{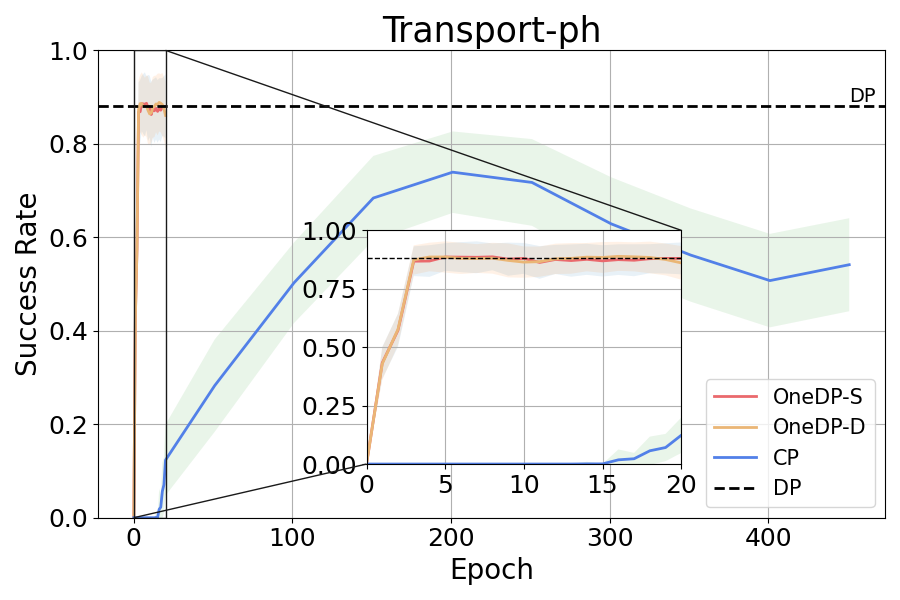}
    \caption{\textbf{Convergence Comparison.} We show our method \ours converges $20\times$ faster than the baseline method Consistency Policy (CP) under EDM setting.}
    \vspace{-5mm}
    \label{fig:convergence}
\end{figure}

\textbf{Experiment Setup.} We pretrain the DP model for 1000 epochs on each benchmark under both DDPM \citep{ho2020denoising} and EDM \citep{karras2022elucidating} noise scheduling. Note EDM noise scheduling is a requirement for CP \citep{prasad2024consistency} to satisfy diffusion boundary conditions. Subsequently, we train \ours for 20 epochs and the baseline CP for 450 epochs until convergence. During evaluation, we observe significant variance in evaluating success rates with different environment initializations. We present average success rates across 5 training seeds and 100 different initial
conditions (500 in total). We report the peak success rate for each method during training, corresponding to the peak points of the curves in \Cref{fig:convergence}.
The metric for most tasks is the success rate, except for PushT, which is evaluated using the coverage of the target area.


Table \ref{tab:main_result_ddpm} presents the results of \ours compared with DP under the default DDPM setting. For DP, we report the average success rate using DDPM sampling with 100 timesteps, as well as the accelerated DDIM sampling with 1 and 10 timesteps. Notably, DP fails to generate reasonable actions with single-step generation, yielding a 0\% success rate for all tasks. DP with 10 steps under DDIM slightly outperforms DP under DDPM. However, \ours demonstrates the highest average success rate with single-step generation across the six tasks, with the stochastic variant \ours-S surpassing the deterministic \ours-D. This superior performance of \ours-S aligns with our discussion in \Cref{sec:diff_distill}, suggesting that stochastic policies generally perform better in complex environments. Interestingly, \ours-S even slightly outperforms the pre-trained DP, which is not unprecedented, as shown in cases of image distillation \citep{zhou2024score} and offline RL \citep{chen2023score}. We attribute this to the fact that iterative sampling may introduce subtle cumulative errors during the denoising process, whereas single-step sampling avoids this issue by jumping directly from the end to the start of the reverse diffusion chain.

In \Cref{tab:main_result_edm}, we report a similar comparison under the EDM setting, including CP. We report DP under the same 1 and 10 DDIM steps, and 100 DDPM steps, which correspond to 1, 19, and 35 number of function evaluations (NFE) in EDM due to second-order ODE sampling. \ours-S outperforms the baseline CP with single-step and its default best setting of 3-step chain generation. Under EDM, \ours-S matches the average success rate of the pre-trained DP, while \ours-D performs slightly worse. We also observe that CP converges much more slowly compared to \ours, as shown in \Cref{fig:convergence}. This slower convergence is likely because CP, based on CTM, does not involve the auxiliary discriminator training that is used to enhance distillation performance in CTM.

\subsection{Real world experiments}
\begin{figure}[ht]
    \centering

\begin{minipage}{\textwidth}
    \centering
    \includegraphics[width=\linewidth]{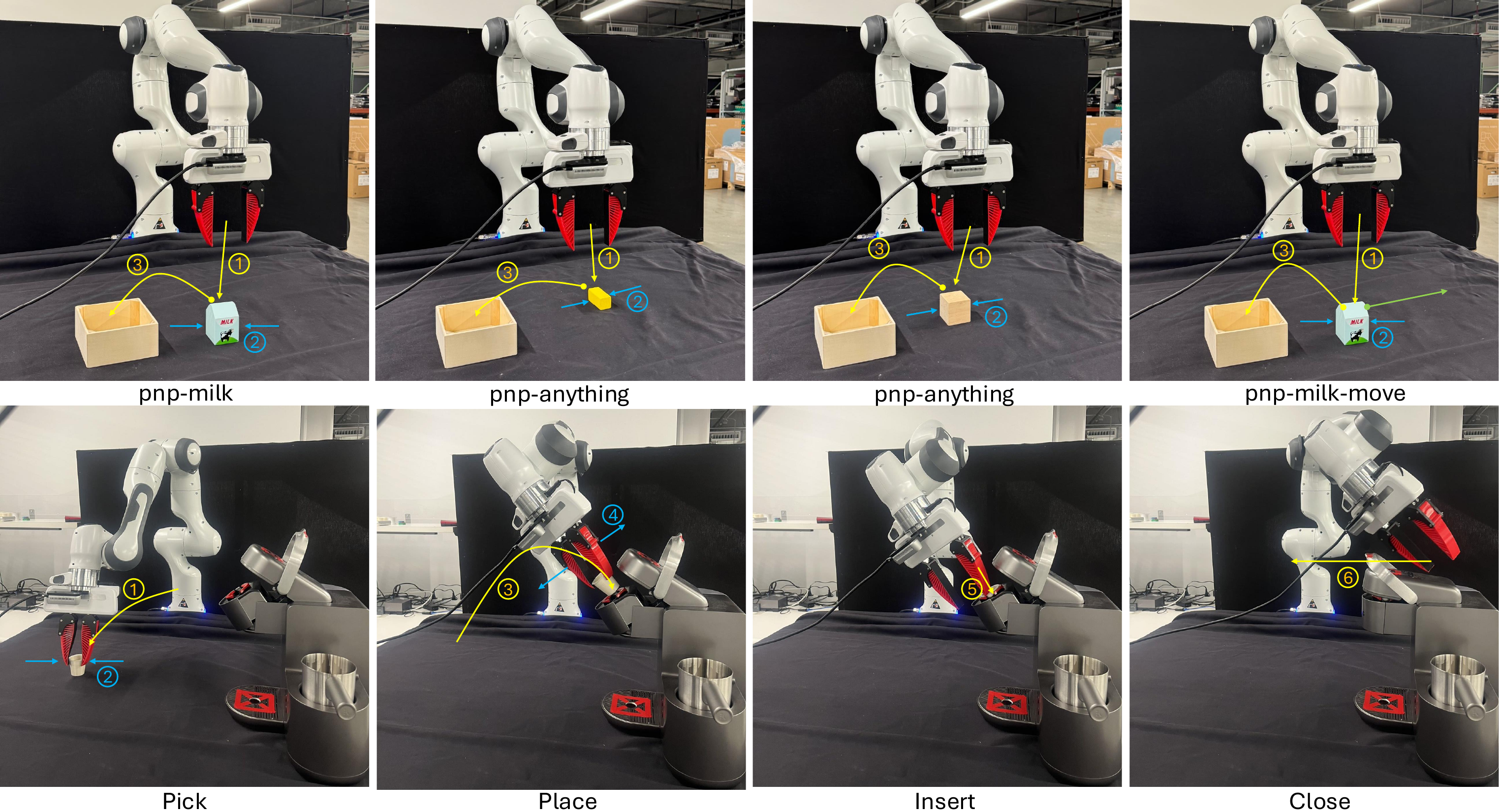}
    \caption{\textbf{Real-World Experiment Illustration.} In the first row, we display the setup for the pick-and-place experiments, featuring three tasks: \texttt{pnp-milk}, \texttt{pnp-anything}, and \texttt{pnp-milk-move}. In total, \texttt{pnp-anything} handles around 10 random objects as shown in \Cref{fig:eval_pnp_anything}. The second row illustrates the procedure for the more challenging \texttt{coffee} task, where the Franka arm is tasked with locating the coffee cup, precisely positioning it in the machine's cup holder, inserting it, and finally closing the machine's lid. }
    \vspace{2mm}
    \label{fig:real_exp_illustration}
\end{minipage}

\begin{minipage}{\textwidth}
\centering
\captionof{table}{\textbf{Success Rate of Real-world Experiments}. We evaluate the performance of our proposed \ours-D and \ours-S against the baseline Diffusion Policy in real-world robotic manipulation tasks. The baseline Diffusion Policy was trained for 1000 epochs to ensure convergence, whereas our distilled models were trained for 100 epochs. We do not select checkpoints; only the final checkpoint is used for evaluation. Performance is assessed over 20 predetermined rounds, and we report the average success rate.}
\label{tab:real_world_success}
\resizebox{0.7\textwidth}{!}{
\begin{tabular}{c|cc|cccc|c}
\hline
\textbf{Method} & \textbf{Epochs} & \textbf{NFE} & \textbf{pnp-milk} & \textbf{pnp-anything} & \textbf{pnp-milk-move} & \textbf{coffee} & \textbf{Avg} \\
\hline
DP(DDIM) & 1000 & 10  & \textbf{1.00} & 0.95 & 0.80 & 0.80 & 0.83 \\
\ours-D & 100 & 1 & \textbf{1.00} & \textbf{1.00} & \textbf{1.00} & 0.80 & 0.95 \\
\ours-S & 100 & 1 & \textbf{1.00} & \textbf{1.00} & \textbf{1.00} & \textbf{0.90} & \textbf{0.98} \\
\hline
\end{tabular}}
\end{minipage}

\begin{minipage}{\textwidth}
\centering
\vspace{2mm}
\captionof{table}{\textbf{Time Efficiency of Real-world Experiments}. We present the completion times for each algorithm as recorded in \Cref{tab:real_world_success}. For a fair comparison, we report the average completion time (in seconds) for each algorithm across evaluation rounds where all algorithms succeeded. Specifically, the tasks \texttt{pnp-milk}, \texttt{pnp-anything}, \texttt{pnp-milk-move}, and \texttt{coffee} were averaged over 18, 15, 8, and 13 respective rounds. These times indicate how quickly each algorithm responds and completes tasks in a real-world environment.}
\label{tab:real_world_time}
\resizebox{0.7\textwidth}{!}{
\begin{tabular}{c|cc|cccc|c}
\hline
\textbf{Method} & \textbf{Epochs} & \textbf{NFE} & \textbf{pnp-milk} & \textbf{pnp-anything} & \textbf{pnp-milk-move} & \textbf{coffee} & \textbf{Avg} \\
\hline
DP(DDIM) & 1000 & 10 & 29.74 & 26.03 & 34.75 & 54.92 & 36.36 \\
\ours-D & 100 & 1 & 23.21 & 22.93 & 28.73 & 33.13 & 27.00 \\
\ours-S & 100 & 1 & \textbf{22.69} & \textbf{22.62} & \textbf{28.15} & \textbf{29.78} & \textbf{25.81} \\
\hline
\end{tabular}}
\end{minipage}

\end{figure}

We design four tasks to evaluate the real-world performance of \ours, including three common tasks where the robot picks and places objects at designated locations, referred to as \texttt{pnp}, and one challenging task where the robot learns to manipulate a coffee machine, called \texttt{coffee}. \Cref{fig:real_exp_illustration} shows the experimental setup, with the first row illustrating the \texttt{pnp} tasks and the second row depicting the \texttt{coffee} task. We introduce the data collection process and the evaluation setup in the following section and provide more details in \Cref{sec:appendix_exp}.

\textbf{\texttt{pnp} Tasks. } This task requires the robot to pick an object from the table and put it in a box. We design three variants of this task: \texttt{pnp-milk}, \texttt{pnp-anything} and \texttt{pnp-milk-move}. In \texttt{pnp-milk}, the object is always the same milk box. In \texttt{pnp-anything}, we expand the target to 11 different objects as shown in \Cref{fig:eval_pnp_anything}. For \texttt{pnp-milk-move}, we involve human interference to create a dynamic environment. Whenever the robot gripper attempts to grasp the milk box, we move it away, following the trajectory as shown in \Cref{fig:eval_pnp_milk_move}. We collect 100 demonstrations each for the \texttt{pnp-milk} and \texttt{pnp-anything} tasks. Separate models are trained for both tasks, with the \texttt{pnp-anything} model utilizing all 200 demonstrations. The \texttt{pnp-milk-move} task is evaluated using the checkpoint from the \texttt{pnp-anything} model.  

\textbf{\texttt{Coffee} Task. } This task requires the robot to operate a coffee machine. It involves the following steps: (1) picking up the coffee pod, (2) placing the coffee pod in the pod holder on the coffee machine, and (3) closing the lid of the coffee machine. This task is more challenging since it involves more steps and requires the robot to insert the pod in the holder accurately. We collect 100 human demonstrations for this task. We train one specific model for this task.

\textbf{Evaluation.} We evaluate the success rate and task completion time from 20 predetermined initial positions for the \texttt{pnp-milk}, \texttt{pnp-anything}, and \texttt{coffee} tasks, as well as 10 motion trajectories for the \texttt{pnp-milk-move} task. The left side of \Cref{fig:time_illustration} shows the setup of the robot, destination box, and coffee machine, with 20 fixed initialization points. \Cref{fig:eval_pnp_milk_move} shows the 10 trajectories for evaluating \texttt{pnp-milk-move}. Details of the evaluation are provided in \Cref{sec:appendix_exp}.
For DP, we follow \citet{chi2023diffusion} to use DDIM (10 steps) to accelerate  the real-world experiment.

We compare \ours against the DP backbone in real-world experiments, focusing on three key aspects: success rate, responsiveness, and time efficiency. \Cref{tab:real_world_success} demonstrates that \ours consistently outperforms DP across all tasks, with the most significant improvement seen in \texttt{pnp-milk-move}. This task demands rapid adaptation to dynamic environmental changes, particularly due to sudden human interference. The wall-clock time for action generation is reported in \Cref{tab:inference_time_real}. The slow action generation of DP hinders its ability to track the moving milk box effectively, often losing control when the box moves out of its visual range, as it is still predicting actions based on outdated information. In contrast, \ours generates actions quickly, allowing it to instantly follow the box's movement, achieving a 100\% success rate in this dynamic task. \ours-S slightly outperforms \ours-D, aligning with the observations from the simulation experiments.

Additionally, we measure the task completion time for successful evaluation rollouts across all algorithms. As shown in \Cref{tab:real_world_time}, \ours completes tasks faster than DP. Both \ours-S and \ours-D exhibit similarly-rapid task completion times. The quick action prediction of \ours reduces hesitation during robot arm movements, particularly when the arm camera's viewpoint changes abruptly. This leads to significant improvements in task completion speed. In \Cref{fig:time_illustration}, we present a heatmap for illustrating the task completion times; lighter colors indicate faster completion times, while dark red demonstrates failure cases. Overall, \ours completes tasks more efficiently across most locations. Although all three algorithms encounter failures in some corner cases for the \texttt{coffee} task, \ours-S shows fewer failures.

\begin{table}[t]
    \centering
    \caption{\textbf{Real-world inference speeds.} We report the wall clock times for each policy in real-world scenarios. The action generation process consists of two parts: observation encoding (OE) and action prediction by each method. All measurements were taken using a local NVIDIA V100 GPU, with the same neural network size for each method. The policy frequencies, shown in \Cref{fig:teaser1}, are based on the values from this table. }
    \resizebox{0.6\textwidth}{!}{
    \begin{tabular}{ccccc}
    \hline
         & OE & DDPM (100 steps) & DDIM (10 steps) & \ours (1 step) \\
    \hline
       Time (ms) & 9 & 660 & 66 & 7\\
       NFE & 1 & 100 & 10 & 1 \\
    \hline
    \end{tabular}}
    \label{tab:inference_time_real}
\end{table}




\section{Related Work} \label{sec:related_work}

\textbf{Diffusion Models. } Diffusion models have emerged as a powerful framework for modeling complex data distributions and have achieved groundbreaking performance across various tasks involving generative modeling \citep{ho2020denoising, karras2022elucidating}. 
They operate by transforming data into Gaussian noise through a diffusion process and subsequently learning to reverse this process via iterative denoising.
Diffusion models have been successfully applied to a wide range of domains, including image, video, and audio generation \cite{saharia2022photorealistic, ramesh2022hierarchical, balaji2022ediff, chen2023videocrafter1, ho2022video, popov2021grad, kong2020diffwave}, reinforcement learning \citep{janner2022planning, wang2022diffusion, psenka2023learning} and robotics \citep{ajay2022conditional, urain2023se, chi2023diffusion}. 

\textbf{Diffusion Policies. }Diffusion models have shown promising results as policy representations for control tasks. 
\citet{janner2022planning} introduced a trajectory-level diffusion model that predicts all timesteps of a plan simultaneously by denoising two-dimensional arrays of state and action pairs. 
\citet{wang2022diffusion} proposed Diffusion Q-learning, which leverages a conditional diffusion model to represent
the policy in offline reinforcement learning. An action-space diffusion model is trained to generate actions conditioned on the states. 
Similarly, \citet{chi2023diffusion} used a conditional diffusion model in the robot action space to represent the visuomotor policy and demonstrated a significant performance boost in imitation learning for various robotics tasks. \citet{ze20243d} further incorporated the power of a compact 3D visual representations to improve diffusion policies in robotics. 

\textbf{Diffusion Distillations. } Although diffusion models are powerful, their iterative denoising process makes them inherently slow in generation, which poses challenges for time-sensitive applications like robotics and real-time control. Motivated by the need to accelerate diffusion models, diffusion distillation has become an active research topic in image generation. Diffusion distillation aims to train a student model that can generate samples with fewer denoising steps by distilling knowledge from a pre-trained teacher model \citep{salimans2022progressive, luhman2021knowledge, zheng2023fast, song2023consistency, kimconsistency}.  \citet{salimans2022progressive} proposed a method to distill a teacher model into a new model that takes half the number of sampling steps, which can be further reduced by progressively applying this procedure. \citet{song2023consistency} introduced consistency models that enable fewer step sampling by enforcing self-consistency of the ODE trajectories. CTM \citep{kimconsistency} improved consistency models and provided the flexibility to trade-off quality and speed. 
\citep{luo2024diff,yin2024one} leverage the success of stochastic distillation sampling \citep{poole2022dreamfusion} in text-to-3D and proposes KL-based score distillation for image generation. Beyond KL, \citet{zhou2024score} proposes the SiD distillation technique derived from Fisher Divergence. 
However, leveraging diffusion distillation to accelerate diffusion policies for robotics remains an underexplored and pressing challenge, particularly for real-time control applications. Consistency Policy \citep{prasad2024consistency} explored applying CTM to reduce the number of denoising steps and accelerate inference of the diffusion policies. It simplifies the original CTM training by ignoring the adversarial auxiliary loss. 
While this approach achieves a considerable speed-up, it leads to performance degradation compared to pre-trained models, and its complex training process and slow convergence present challenges for robotics applications. 
In contrast, \ours employs expectational reverse KL optimization to distill a powerful one-step action generator, achieving comparable or higher success rates than the original diffusion policy, while converging 20$\times$ faster.



\section{Conclusion}

In this paper, we introduced the One-Step Diffusion Policy (\ours) through advanced diffusion distillation techniques. We enhanced the slow, iterative action prediction process of Diffusion Policy by reducing it to a single-step process, dramatically decreasing action inference time and enabling the robot to respond quickly to environmental changes. Through extensive simulation and real-world experiments, we demonstrate that \ours not only achieves a slightly higher success rate, but also responds quickly and effectively to environmental interference. The rapid action prediction further allows the robot to complete tasks more efficiently.

However, this work has some limitations. In the experiments, we did not test \ours on long-horizon real-world tasks. Furthermore, in the real-world experiments, we limited the robot’s operation frequency to 20 Hz for controlling stability, which underutilized \ours’s full potential. Additionally, the KL-based distillation method may not be the optimal choice for distribution matching, and introducing a discriminator term could potentially improve distillation performance. 

\bibliography{iclr2025_conference}
\bibliographystyle{iclr2025_conference}

\newpage
\appendix
\section{Real-World Experiment Setup} \label{sec:appendix_exp}

\begin{figure}[h]
    \centering
    \includegraphics[width=0.5\linewidth]{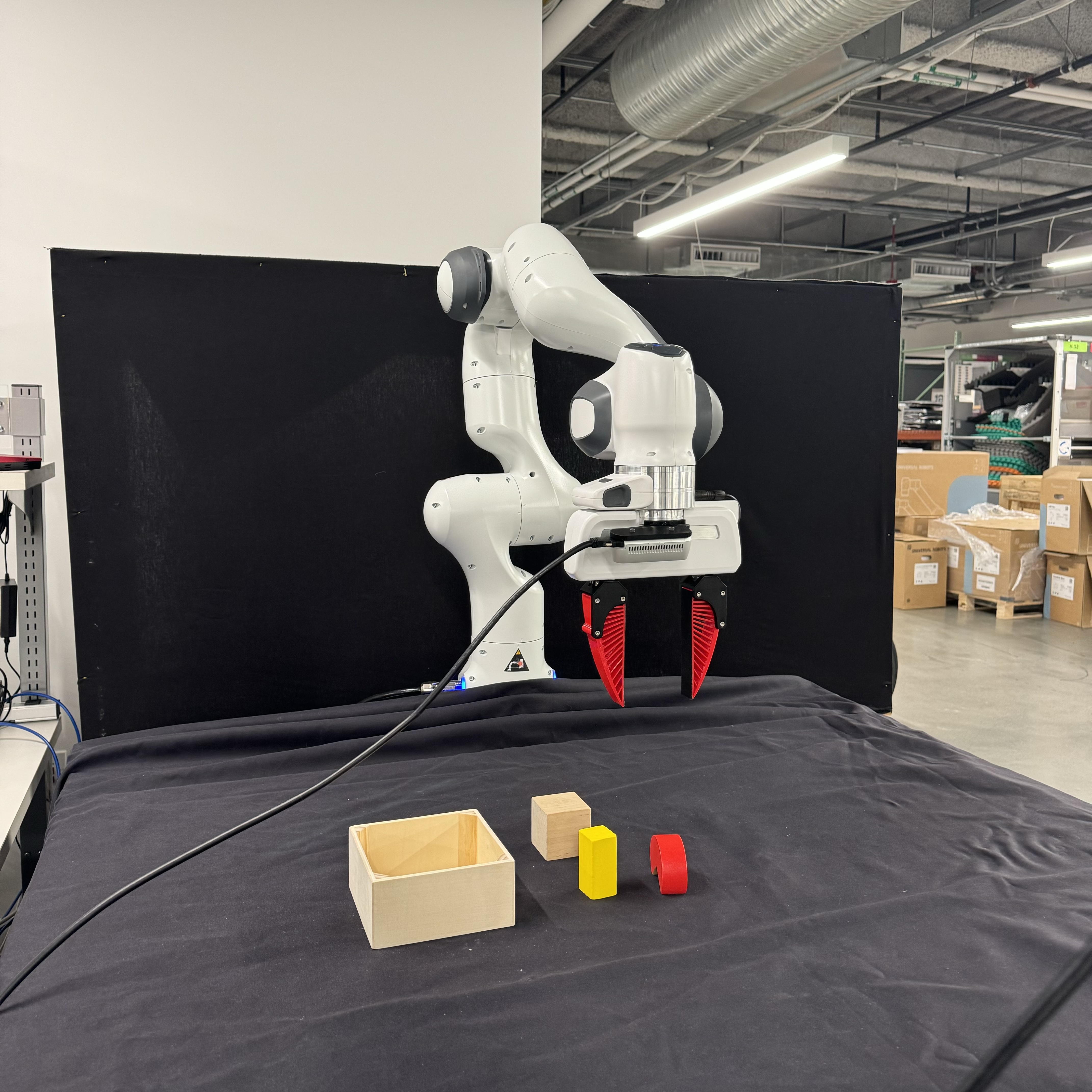}
    \caption{Real-world Experiment Setup}
    \label{fig:real_setup}
\end{figure}

\paragraph{Robot Setup.} The physical robot setup consists of a Franka Panda robot arm, a front-view Intel RealSense D415 RGB-D camera, and a wrist-mounted Intel RealSense D435 RGB-D camera. The RGB image resolution was set to 120x160. The depth image is not used in our experiments. 

\paragraph{Teleoperation.} Demonstration data for the real robot tasks was collected using a phone-based teleoperation system~\citep{mandlekar2018roboturk, mandlekar2019scaling}.

\begin{figure}[t]
    \centering
    \includegraphics[width=\linewidth]{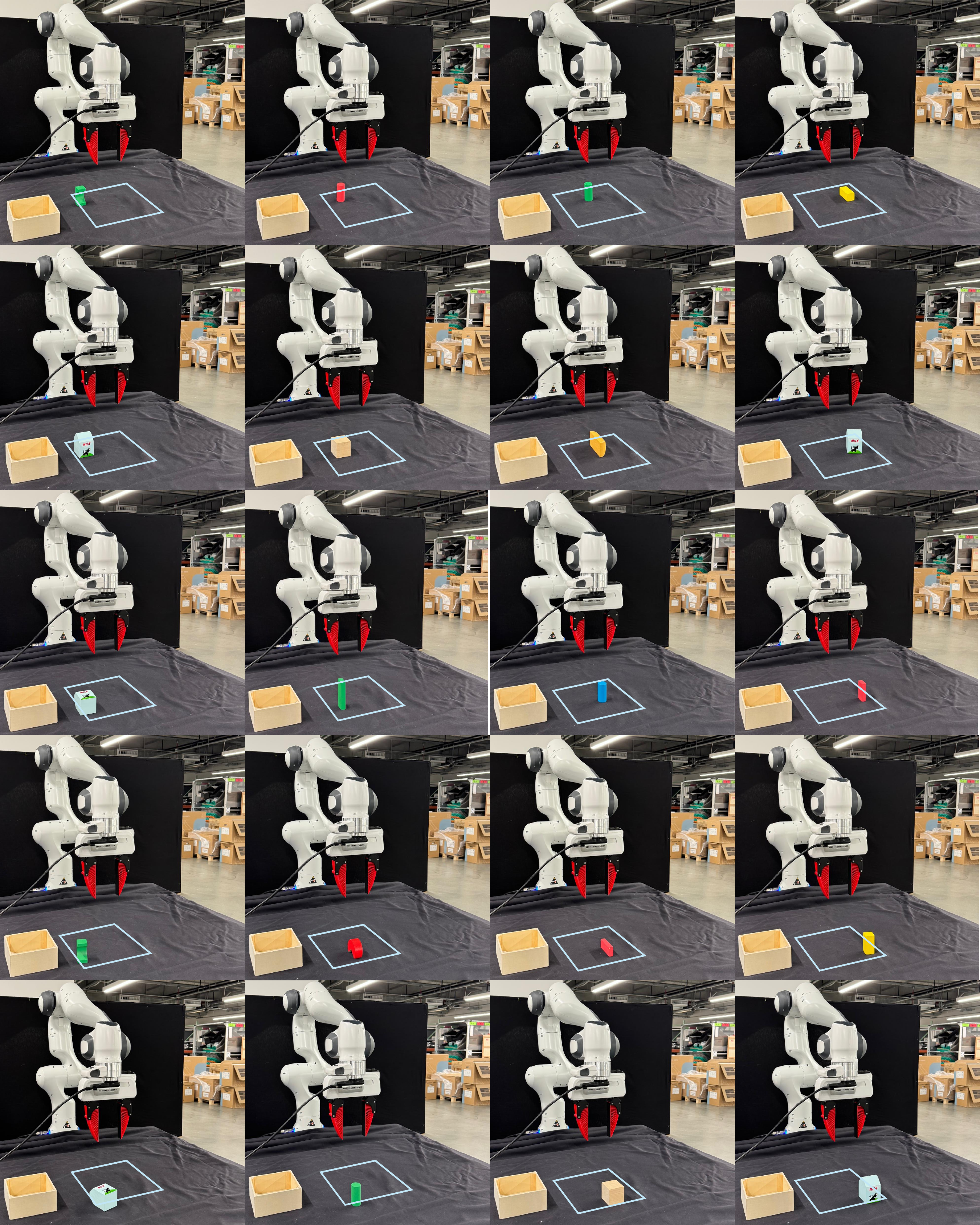}
    \caption{Evaluation setup for \texttt{pnp-anything}.}
    \label{fig:eval_pnp_anything}
\end{figure}

\paragraph{Data Collection.} We collect 100 demonstrations for each task separately: \texttt{pnp-milk}, \texttt{pnp-anything}, and \texttt{coffee}. In \texttt{pnp-milk}, the target object is always the milk box, and the task involves picking up the milk box from various random locations and placing it into a designated target box at a fixed location. For \texttt{pnp-anything}, we extend the set of target objects to 11 different items, as shown in \Cref{fig:eval_pnp_anything}, with the target box location randomized vertically. In the \texttt{coffee} task, the coffee cup is randomly placed, and the robot is required to pick it up, insert it into the coffee machine, and close the lid.

\begin{figure}[t]
    \centering
    \includegraphics[width=\linewidth]{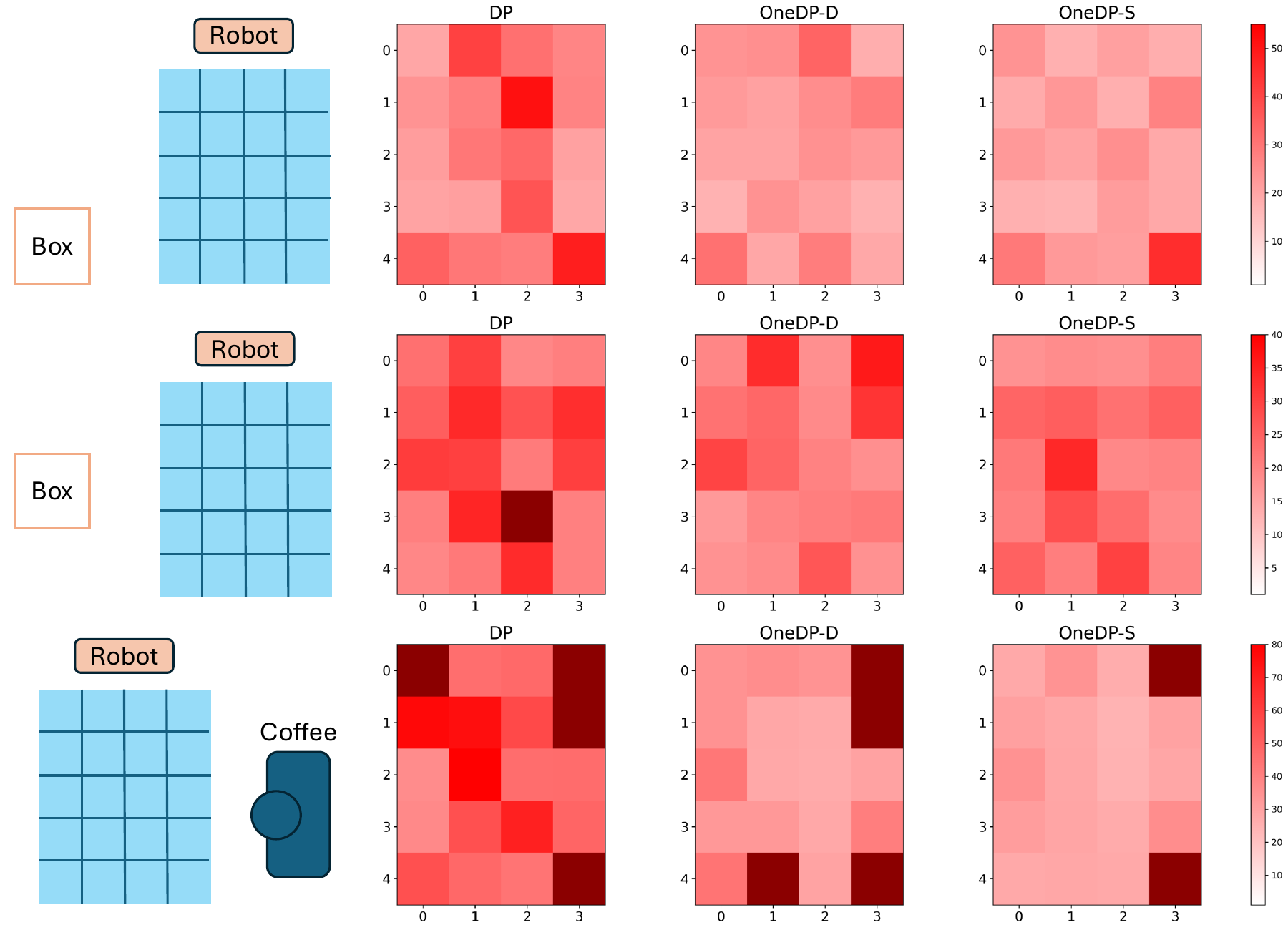}
    \caption{\textbf{Real-World Comparison Illustration.} We present the time taken by each algorithm to complete tasks from a specific starting point in colors. A color map on the right side ranges from white to red indicating the time in seconds. Dark red signifies that the algorithm failed at that location. The three rows represent tasks \texttt{pnp-milk}, \texttt{pnp-anything}, \texttt{coffee}. Details of the evaluation of \texttt{pnp-anything} can be found in \Cref{fig:eval_pnp_anything}.}
    \label{fig:time_illustration}
\end{figure}

The area and location for each task are illustrated in the left column of \Cref{fig:time_illustration}. During data collection, target objects are randomly positioned within the blue area; the grid is used for evaluation, as described in the next section.
For the \texttt{pnp} tasks, the blue area is a rectangle measuring 23 cm in height and 20 cm in width, while the target box is a square with a side length of 13 cm. In the \texttt{coffee} task, the blue area is slightly smaller, measuring 18 cm in height and 20 cm in width.

\begin{table}[h]
    \centering
    \caption{\textbf{Real-world experiment demonstrations.} In total we collect 300 demonstrations, with 100 demonstrations for each task. }
    \resizebox{0.5\textwidth}{!}{
    \begin{tabular}{cccc}
    \hline
         & pnp-milk & pnp-anything & coffee \\
    \hline
       Demos & 100 & 100 & 100 \\
    \hline
    \end{tabular}}
    \label{tab:demonstraions}
\end{table}

\begin{figure}[t]
    \centering
    \includegraphics[width=\linewidth]{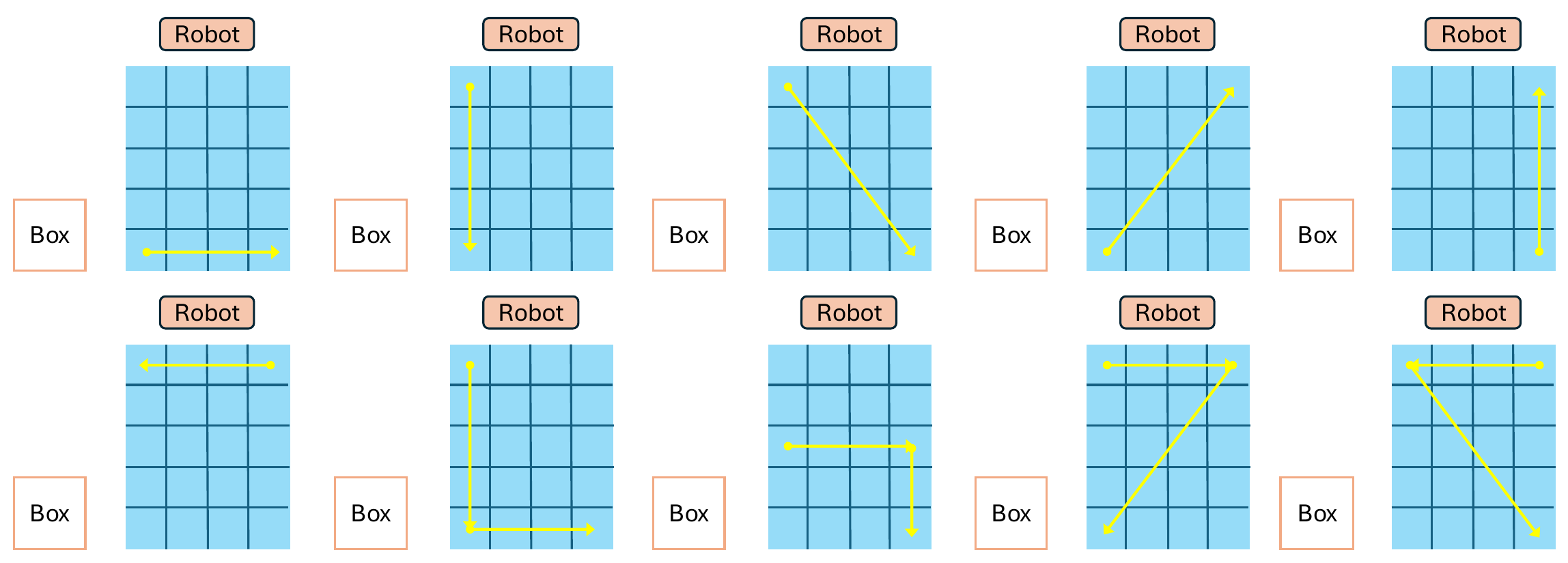}
    \caption{Evaluation trajectories for \texttt{pnp-milk-move}. The box is always on the left-hand side of the tested blue area.}
    \label{fig:eval_pnp_milk_move}
\end{figure}

\paragraph{Evaluation.} To ensure a fair comparison between OneDP and all baseline methods, we standardize the evaluation process. For the \texttt{pnp-milk}, \texttt{pnp-anything}, and \texttt{coffee} tasks, we evaluate each method according to the grid order shown in \Cref{fig:time_illustration}. The target object is placed at the center of the grid to ensure consistent initial conditions across evaluations. For task \texttt{pnp-anything}, the picked object also follows the order shown in \Cref{fig:eval_pnp_anything}.
For the dynamic environment task \texttt{pnp-milk-move}, we introduce human interference during the evaluation. Whenever the robot gripper attempts to grasp the target milk box, we manually move it away along the trajectory depicted in \Cref{fig:eval_pnp_milk_move}. Although we aim to maintain consistent conditions during each evaluation, the exact nature of human interference cannot be guaranteed. Some trajectories involve a single instance of interference, while others may involve two consecutive human movements. 

The original DDPM sampling in Diffusion Policy is too slow for real-world experiments. To speed up the evaluation, we follow \citep{chi2023diffusion} and use DDIM with 10 steps. For OneDP, we use single-step generation. In real-world experiments, we do not select intermediate checkpoints but use the final checkpoint after training for each method.

We record both the success rates and completion times, reporting their mean values. For \texttt{pnp-milk-move}, evaluations are conducted over 10 trajectories, while for the other tasks, results are obtained from 20 grid points. In \Cref{fig:time_illustration}, we present a heatmap to visualize task completion times, where lighter colors represent faster completions and dark red indicates failure cases. Overall, \ours completes tasks more efficiently across most locations. While all three algorithms experience failures in certain corner cases for the \texttt{coffee} task, \ours-S demonstrates fewer failures.

\section{Training Details}

We follow the CNN-based neural network architecture and observation encoder design from \citet{chi2023diffusion}. For simulation experiments, we use a 256-million-parameter version for DDPM and a 67-million-parameter version for EDM, as the smaller EDM network performs slightly better. In real-world experiments, we also use the 67-million-parameter version. Additionally, we adopt the action chunking idea from \citet{chi2023diffusion} and \citet{zhao2023learning}, using 16 actions per chunk for prediction, and utilize two observations for vision encoding.

We first train DP for 1000 epochs in both simulation and real-world experiments with a default learning rate of 1e-4 and weight decay of 1e-6. We then perform distillation using the pre-trained checkpoints, distilling for 20 epochs in simulation and 100 epochs in real-world experiments.

For distillation, we warm-start both the stochastic and deterministic action generators, \( G_\theta \), and the generator score network, \( \epsilon_\psi \), by duplicating the network structure and weights from the pre-trained diffusion-policy checkpoints. Since the generator network is initialized from a denoising network, a timestep input is required, as this was part of the original input. We fix the timestep at 65 for discrete diffusion and choose \( \sigma = 2.5 \) for continuous EDM diffusion. The generator learning rate is set to 1e-6. We find these hyperparameters to be stable without causing significant performance variation. We provide an ablation study that focuses primarily on the generator score network’s learning rate and optimizer settings in \Cref{sec:exp_ablation}. We provide the hyperparameter details in \Cref{tab:hyperparam}.

\begin{table}[h]
    \centering
    \begin{tabular}{cc}
        \hline
        Hyperparameters & Values \\
        \hline
        generator learning rate & lr=1e-6 \\
        generator score network learning rate & lr=2e-5 \\
        generator optimizer & Adam([0.0, 0.999]) \\
        generator score network optimizer & Adam([0.0, 0.999]) \\
        action chunk size & n=16 \\
        number of observations & n=2 \\
        discrete diffusion init timestep & $t_{\text{init}}$=65 \\
        discrete diffusion distillation $t$ range & [2, 95] \\
        continuous diffusion init sigma & $\sigma=2.5$ \\
        \hline
    \end{tabular}
    \caption{Hyperparameters}
    \label{tab:hyperparam}
\end{table}

\section{Ablation Study} \label{sec:exp_ablation}

\begin{figure}
    \centering
    \includegraphics[width=0.48\linewidth]{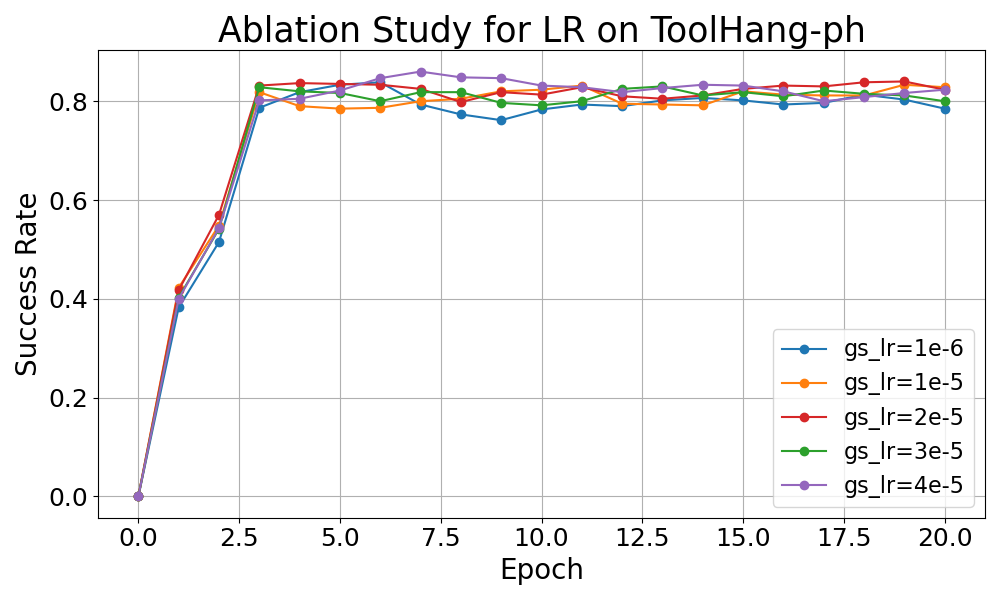}
    \includegraphics[width=0.48\linewidth]{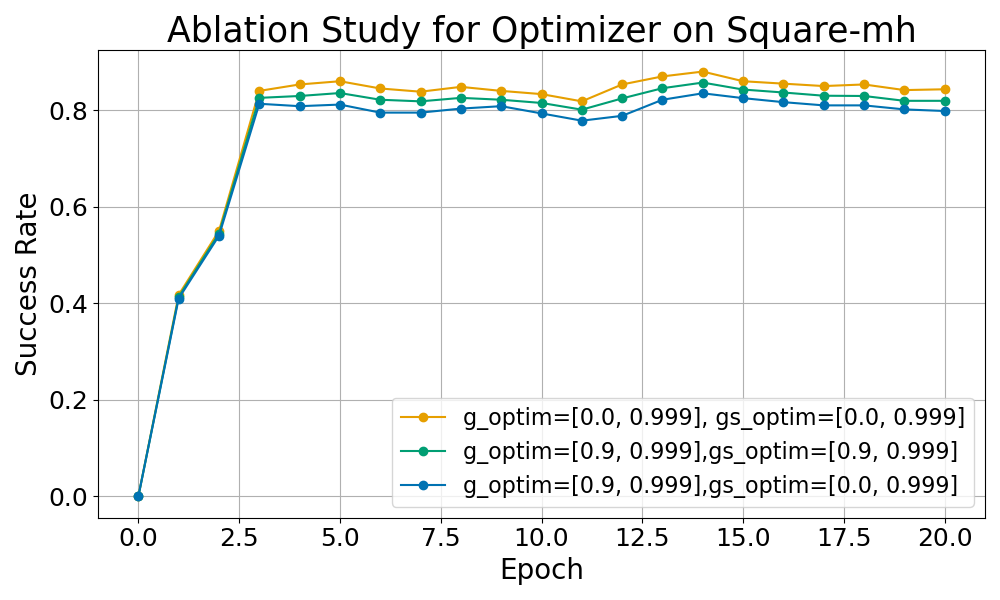}
    \caption{Ablation studies on the learning rate of the generator score network and optimizer settings. }
    \label{fig:ablation}
\end{figure}

As shown in the first panel of \Cref{fig:ablation}, we explore a range of learning rates for the generator score network in the grid [1e-6, 1e-5, 2e-5, 3e-5, 4e-5] and find 2e-5 to be optimal in most cases. A higher learning rate for the score network compared to the generator ensures that the score network keeps pace with the generator's distribution updates during training. In the second panel, we search for the best optimizer settings, finding that setting \(\beta_1\) to 0 for both the generator and the generator score network optimizers is effective. This approach, commonly used in GANs, allows the two networks to evolve together more quickly.

\section{Detailed Preliminaries} \label{sec:appendix_prelim}

Diffusion models are robust generative models utilized across various domains \citep{ho2020denoising, sohl2015deep, song2020score}. They operate by establishing a forward diffusion process that incrementally transforms the data distribution into a known noise distribution, such as standard Gaussian noise. A probabilistic model is then trained to methodically reverse this diffusion process, enabling the generation of data samples from pure noise.

Suppose the data distribution is $p(\vx)$. The forward diffusion process is conducted by gradually adding Gaussian noise to samples $\vx^0 \sim p(\vx)$ as follows,
\begin{equation*}
    \vx^k = \alpha_k \vx^0 + \sigma_k \vepsilon_k, \vepsilon_k \sim \gN(\bm{0}, \mI); \quad q(\vx^k | \vx^0) := \gN(\alpha_k \vx^0, \sigma_k^2 \mI)
\end{equation*}
where $\alpha_k$ and $\sigma_k$ are parameters manually designed to vary according to different noise scheduling strategies. DDPM \citep{ho2020denoising} is a discrete-time diffusion model with $k \in \{1,\dots,K\}$. It can be easily extended to continuous-time diffusion from the score-based generative model perspective \citep{song2020score, karras2022elucidating} with $k \in [0,1]$. With sufficient amount of noise added, $\vx^K \simeq \gN(\bm{0}, \mI)$. 
\citet{ho2020denoising} propose to reverse the diffusion process and iteratively reconstruct the original sample $\vx^0$ by training a neural network $\epsilon_{\theta}(\vx^k, k)$ to predict the noise $\vepsilon_k$ added at each forward diffusion step (epsilon prediction). 
With reparameterization $\vepsilon_k = (\vx^k - \alpha_k \vx^0) / \sigma_k$, the diffusion model could also be formulated as a $\vx^0$-prediction process $x_\theta(\vx^k, k)$ \citep{karras2022elucidating, xiao2021tackling}. We use epsilon prediction $\epsilon_\theta$ in our derivation. The diffusion model is trained with the denoising score matching loss \citep{ho2020denoising},
\begin{equation*}
    \min_\theta \E_{\vx^k \sim q(\vx^k|\vx^0), \vx^0 \sim p(\vx), k\sim \gU}[\lambda(k) \cdot || \epsilon_\theta(\vx^k, k) - \vepsilon_k||^2]
\end{equation*}
where $\gU$ is a uniform distribution over the $k$ space, and $\lambda(k)$ is a noise-ratio re-weighting function. With a trained diffusion model, we could sample $\vx^0$ by reversing the diffusion chain, which involves discretizing the ODE \citep{song2020score} as follows:
\begin{equation} \label{eq:solve_dm_ode}
    d\vx^{k} = \left[ f(k) \vx^k - \frac{1}{2}g^2(k) \nabla_{\vx_k} \log q(\vx^k ) \right] dk
\end{equation}
where $f(k) = \frac{d \log \alpha_k}{dk}$ and $g^2(k)=\frac{d\sigma_k^2}{dk} - 2\frac{d \log \alpha_k}{dk} \sigma_k^2$. The unknown score $\nabla_{\vx_k} \log q(\vx^k)$ could be estimated as follows:
\begin{equation*}
    s(\vx^k) = \nabla_{\vx_k} \log q(\vx^k) = - \frac{\epsilon^*(\vx^k, k)}{\sigma_k} \approx - \frac{\epsilon_\theta(\vx^k, k)}{\sigma_k},
\end{equation*}
where $\epsilon^*(\vx^k, k)$ is the true noise added at time $k$, and we let $s_\theta(\vx^k) = - \frac{\epsilon_\theta(\vx^k, k)}{\sigma_k} $.

\citet{wang2022diffusion, chi2023diffusion} extend diffusion models as expressive and powerful policies for offline RL and robotics. In robotics, a set of past observation images $\rmO$ is used as input to the policy. An action chunk $\rmA$, which consists of a sequence of consecutive actions, forms the output of the policy. ResNet \citep{he2016deep} based vision encoders are commonly utilized to encode multiple camera observation images into observation features. Diffusion policy is represented as a conditional diffusion-based action prediction model, 
\begin{equation*}
    \pi_\theta(\rmA^0_t | \rmO_t) := \int \cdots \int \gN(\rmA^K_t; \bm{0}, \mI) \prod_{k=K}^{k=1} p_{\theta}(\rmA^{k-1}_t| \rmA^k_t, \rmO_t) d\rmA_t^{K} \cdots d\rmA_t^{1},
\end{equation*}
where $\rmO_t$ contains the current and a few previous vision observation features at timestep $t$, and $p_\theta$ could be represented by $\epsilon_\theta$ as shown in DDPM \citep{ho2020denoising}. The explicit form of $\pi_\theta(\rmA^0_t | \rmO_t)$ is often impractical due to the complexity of integrating actions from $\rmA_t^K$ to $\rmA_t^1$. However, we can obtain an action chunk prediction $\rmA_t^0$ by iteratively solving \Cref{eq:solve_dm_ode} from $K$ to $0$.



\end{document}